\PassOptionsToPackage{table}{xcolor}

\documentclass[10pt,twocolumn,letterpaper]{article}

\usepackage[pagenumbers]{cvpr} 

\definecolor{cvprblue}{rgb}{0.21,0.49,0.74}
\usepackage[pagebackref,breaklinks,colorlinks,allcolors=cvprblue]{hyperref}

\definecolor{lightgray}{gray}{0.95}

\usepackage{multirow}

\usepackage{bbm}

\usepackage[accsupp]{axessibility}

\def\paperID{3282} 
\def\confName{CVPR}
\def\confYear{2025}

\title{Parametric Point Cloud Completion for Polygonal Surface Reconstruction}

\newcommand{\affiliationFont}{\fontsize{11pt}{12pt}\selectfont}
\author{ Zhaiyu Chen$^{1,3}$ \quad Yuqing Wang$^1$ \quad Liangliang Nan$^2$ \quad Xiao Xiang Zhu$^{1,3}$\\[0.15cm]
  {\affiliationFont
    $^{1}$Technical University of Munich \quad
    $^{2}$Delft University of Technology \quad
    $^{3}$Munich Center for Machine Learning
  }
}

\begin{document}
\maketitle
\begin{abstract}
Existing polygonal surface reconstruction methods heavily depend on input completeness and struggle with incomplete point clouds. We argue that while current point cloud completion techniques may recover missing points, they are not optimized for polygonal surface reconstruction, where the parametric representation of underlying surfaces remains overlooked. To address this gap, we introduce parametric completion, a novel paradigm for point cloud completion, which recovers parametric primitives instead of individual points to convey high-level geometric structures. Our presented approach, PaCo, enables high-quality polygonal surface reconstruction by leveraging plane proxies that encapsulate both plane parameters and inlier points, proving particularly effective in challenging scenarios with highly incomplete data. Comprehensive evaluations of our approach on the ABC dataset establish its effectiveness with superior performance and set a new standard for polygonal surface reconstruction from incomplete data. Project page: \url{https://parametric-completion.github.io}.
\end{abstract}
    
\section{Introduction}
\label{sec:intro}

Unlike conventional methods that produce dense triangle meshes~\cite{kazhdan2013screened,erler2020points2surf,huang2024surface}, polygonal surface reconstruction prioritizes a compact surface representation~\cite{nan2017polyfit,bauchet2020kinetic,sulzer2024concise}. The compact polygonal representation is valuable for applications where efficiency, interpretability, and editability are essential~\cite{botsch2010polygon}. However, recovering accurate polygonal surfaces from incomplete point clouds remains a challenge due to factors such as occlusions and sensor limitations. This demands methods that can infer structured, high-level surface representations (\eg, planes) to effectively convey underlying geometries.

\begin{figure}[ht]
  \centering
  \includegraphics[width=0.99\linewidth]{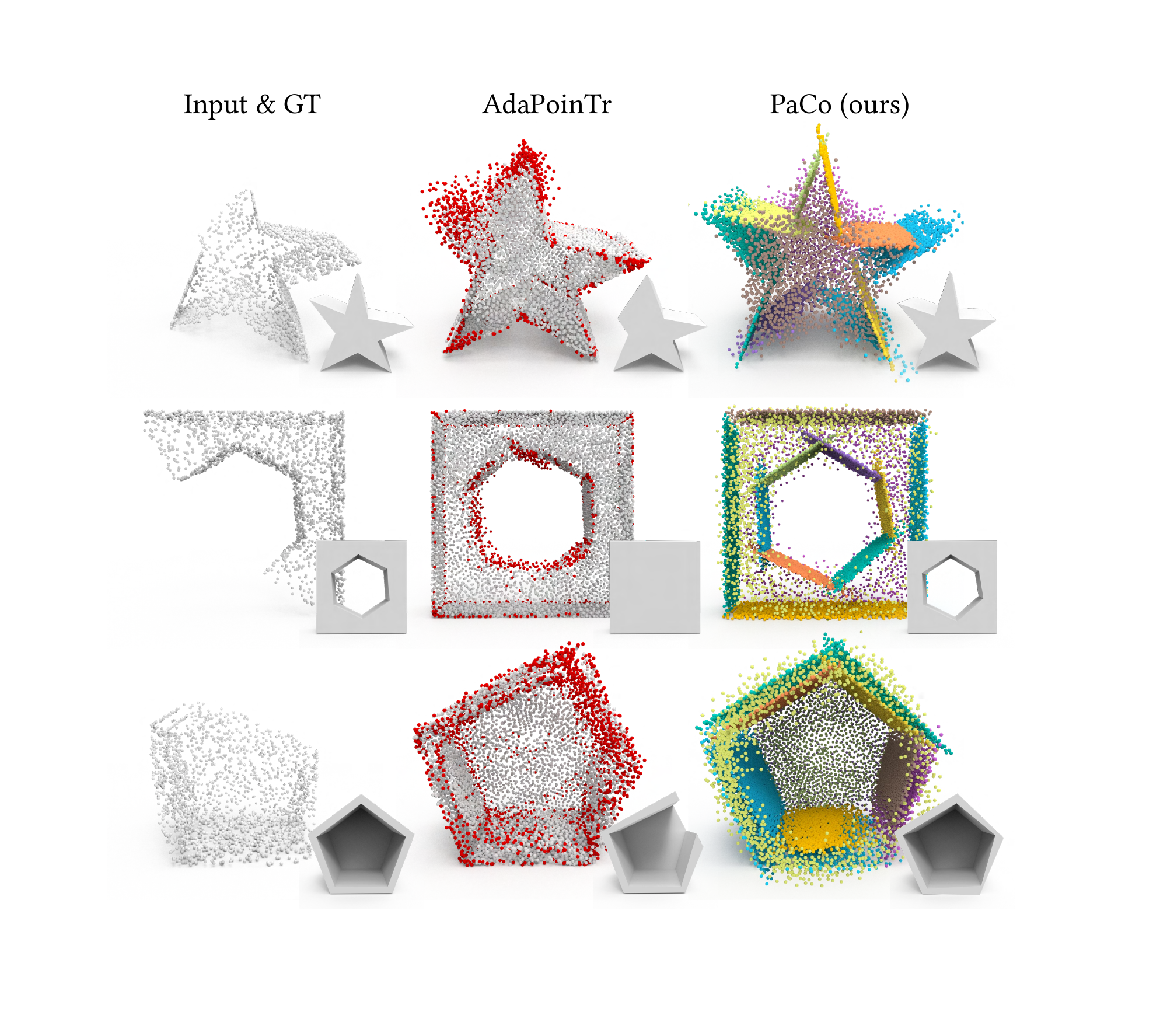}
  \caption{Unlike conventional point cloud completion that aims to recover individual points (\textit{middle}, non-planar points shown in red), parametric completion recovers primitive parameters along with corresponding inlier points. PaCo recovers planar structures (\textit{right}, each color denoting a planar primitive), \textit{directly} enabling high-quality polygonal surface reconstruction.}
  \label{fig:teaser}
\end{figure}

One might consider point cloud completion as a potential solution. However, current practices in point cloud completion aim to recover individual missing points~\cite{yuan2018pcn,yu2023adapointr,yu2021pointr}. While existing methods show promise for free-form object completion, they typically fail to produce structured reconstructions with piecewise planar geometry. The emphasis on point completeness in existing methods overlooks the need for a parametric planar representation, which is essential for accurately reconstructing polygonal surfaces. We argue that a persistent gap exists between the goals of point cloud completion and polygonal surface reconstruction. To our knowledge, none of the existing completion techniques are optimized for polygonal surfaces.

To bridge this gap, we introduce \textit{\textbf{parametric point cloud completion, a novel point cloud completion paradigm}}. Unlike conventional methods that emphasize individual point recovery to improve completeness, parametric completion aims to infer parametric primitives from incomplete data. This shift prioritizes the recovery of high-level geometric structures over point completeness. We present PaCo, a neural network that embodies this paradigm and leverages sequence-to-sequence generation to learn plane proxies. The learned proxies convey the underlying structures of complete surfaces from incomplete inputs, encoding both plane parameters and corresponding inlier points. We then formulate a bipartite matching framework with multiple objectives to optimize primitive distribution. By leveraging these parametric primitives, PaCo effectively enables the reconstruction of high-quality polygonal surfaces from incomplete data, even in challenging scenarios where conventional completion methods typically fail. \cref{fig:teaser} highlights the distinct advantages of PaCo over existing methods.

Benefiting from our parametric representation and associated learning strategies, our approach bridges the gap between shape completion and polygonal surface reconstruction. We summarize our contributions as follows:
\begin{itemize}
    \item We introduce parametric point cloud completion, a new completion paradigm aimed at inferring parametric primitives that convey the underlying structures of object surfaces from incomplete input data.
    \item We present PaCo, a novel deep neural network that embodies the parametric completion paradigm for polygonal surfaces. 
    \item We develop a bipartite matching framework with multiple learning objectives that optimally distributes parametric primitives for accurate and structured completion. 
    \item We establish the effectiveness of parametric completion for reconstructing polygonal surfaces from incomplete point clouds, setting a new standard for this task.

\end{itemize}

\section{Related Work}
\label{sec:relatedwork}


\paragraph{3D Shape Completion.}

Early 3D shape completion methods often rely on volumetric grid data structures~\cite{wu2015, dai2017, xie2020grnet}, applying convolutions to process the data. However, converting point clouds into voxels introduces discretization errors and incurs high computational costs at fine resolutions. To enable direct operation on raw point sets, PointNet~\cite{qi2017pointnet} utilizes symmetric functions to achieve permutation invariance. This breakthrough led to the development of various point-based neural networks~\cite{qi2017pointnet++, wang2019dynamic, thomas2019kpconv} that efficiently consume point cloud data. Point-based methods have since become the mainstream for point cloud completion, with models like PCN~\cite{yuan2018pcn}, FoldingNet~\cite{yang2018foldingnet}, PoinTr~\cite{yu2021pointr}, AdaPoinTr~\cite{yu2023adapointr}, ODGNet~\cite{cai2024orthogonal}, and many more~\cite{xiang2021snowflakenet, zhao2022seedformer, tang2022lake, yan2022fbnet, li2023proxyformer, zhu2023csdn, tesema2023point} demonstrating effectiveness in filling missing points. However, these methods often overlook the underlying geometric structures of surfaces while focusing on individual point recovery. In contrast, our approach emphasizes recovering a parametric planar representation that captures higher-level geometric structures essential for polygonal surfaces.

\paragraph{Polygonal Surface Reconstruction.}

Polygonal surfaces can be reconstructed using primitive assembly methods, which optimize the arrangement of planar primitives or their equivalents. Key examples include PolyFit~\cite{nan2017polyfit}, KSR~\cite{bauchet2020kinetic}, COMPOD~\cite{sulzer2024concise}, and domain-specific ones~\cite{chen2022points2poly, huang2022city3d, chen2024polygnn}. These methods rely on high-quality primitives, often extracted using techniques such as RANSAC~\cite{schnabel2007efficient} and GoCoPP~\cite{yu2022finding}. An alternative approach involves geometric simplification, which reduces mesh complexity while retaining key geometric features~\cite{garland1997qem, salinas2015samd, cohen2004vsa, li2021simplification, gao2022lowpoly, chen2023robust}. Geometric simplification typically begins with a dense mesh generated by methods like Poisson~\cite{kazhdan2013screened} or neural solvers~\cite{erler2020points2surf, boulch2022poco, huang2022neural, huang2023neural}. However, geometric simplification requires high-quality input meshes and typically underperforms compared to primitive assembly to achieve accurate approximations. Notably, neural methods have also shown potential in reconstructing parametric surface meshes~\cite{chen2020bspnet, ren2021csg, yu2022capri, li2023secad, liu2024point2cad} but struggle with incomplete data, which are common due to occlusions and sensor limitations~\cite{huang2024surface}. These limitations motivate our approach: deriving a parametric representation directly from incomplete point clouds to facilitate robust polygonal surface reconstruction via primitive assembly.

\section{Method}
\label{sec:method}

\begin{figure*}[ht]
  \centering
  \includegraphics[width=0.99\linewidth]{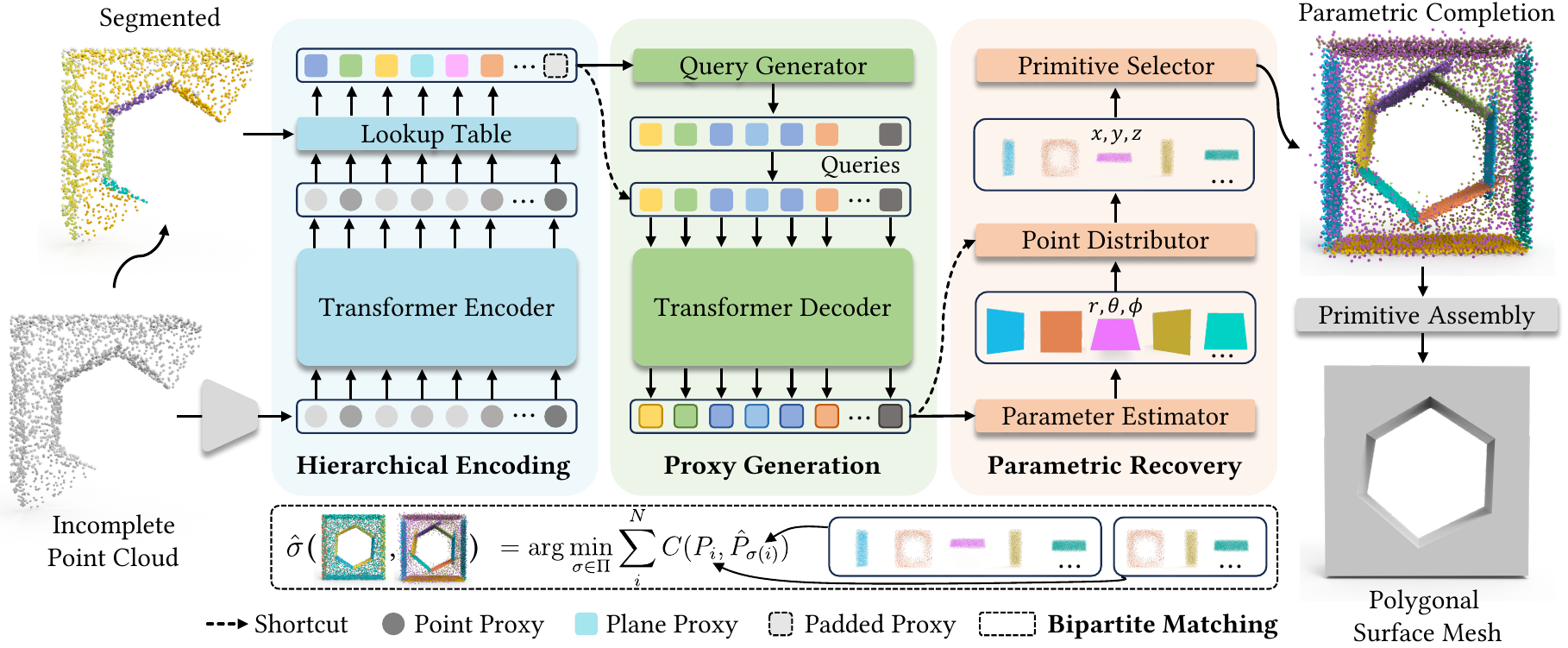}
  \caption{\textbf{Architecture of PaCo.} Starting from an incomplete point cloud, we hierarchically encode the points into plane proxies with a lookup table from segmentation. These proxies inform the proxy generator in producing a set of proxy proposals. The generated proxies are then passed to the parameter estimator and the point distributor for primitive estimation. Finally, the primitive selector identifies a subset of primitives to form the parametric completion, with selection optimized via bipartite matching. The resulting primitives enable \textit{direct} assembly for polygonal surface reconstruction. Points are colored according to their planar primitives.}
  \label{fig:architecture}
\end{figure*}

Given an incomplete point cloud \( X \), our objective is to infer \textit{all} parametric planar primitives \( \hat{P} \) that constitute the complete surface of the object, with each primitive representing a piecewise planar region. We start by hierarchically encoding the input points into plane proxies \( V \) that represent the planar primitives in the input. These proxies are then fed to a proxy generator to produce a set of proxy proposals for the entire surface. Next, a parametric recovery module extracts the plane parameters and distributes their inlier points from these proxies. Unlike existing completion methods that predict a fixed number of points, PaCo uses variable-size primitives that adapt to surfaces of different complexities. This is achieved through bipartite matching during training, while during inference, primitives are selected based on confidence scores from a primitive selector. \cref{fig:architecture} presents the overall architecture of PaCo.

\subsection{Hierarchical Encoding}

Given \(X=\{x_i\}_{i=1}^{N_x}\), we first group the points into a set of planar segments \( S = \{s_i\}_{i=1}^{N_s} \) using GoCoPP~\cite{yu2022finding}, represented by a surjective mapping \( f: X \rightarrow S \). Our encoder progressively builds features across three topological levels: point, point patch, and plane, guided by this mapping.

To start, the points are encoded as point proxies \(X'\) following the PoinTr encoding scheme~\cite{yu2021pointr}. Instead of encoding individual points, these point proxies represent features of local point patches, with each patch’s location defined by its center point in the incomplete point cloud. Since the geometry of \(X\) covers \(X'\), we establish a lossless mapping \( f': X \rightarrow X' \rightarrow S \) using a lookup table derived from the mapping \(f\), allowing hierarchical feature aggregation to the plane level. Finally, we inject each plane's normal embedding into the plane-level semantic feature, incorporating structural information into the plane proxies \(V=\{v_i\}_{i=1}^{K}\). Thus, the \(i\)-th plane proxy \( v_i \) is computed as
\begin{equation}
    v_i = \textit{sum}(X_{i}') + \Phi(n_i),
\end{equation}
where \( \textit{sum} \) denotes aggregation with sum pooling, \( X_{i}' \subseteq X' \) represents all point proxies on the plane, \( \Phi \) is an MLP, and \(n_i\) represents the normal of the plane.

\subsection{Proxy Generation}
The proxy generation module takes the input plane proxies and generates a set of plane proxy proposals, which complete the surface of the object by filling in planes that are partially missing or completely absent in the input.

Since \(V\) varies in length, we first pad it to a fixed size and apply an MLP to produce a set of queries \( Q_I \) representing primitives in the input. \( V \)  is then passed to a query generator, producing another set of queries \( Q_G \) representing the missing primitives. We adopt the query ranking strategy~\cite{yu2023adapointr} to select the top \( M \) queries from both sources combined. A Transformer block~\cite{vaswani2017attention} then attends to the input plane proxies \(V\) and the selected queries \(Q=\{q_i\}_{i=1}^{M}\) to generate a set of plane proxies as proposals.

\subsection{Parametric Recovery}
The generated proxies are fed into the parametric recovery module to retrieve plane parameters and inlier points, using three prediction heads: parameter estimator, point distributor, and primitive selector.

\paragraph{Parameter Estimator.}
The parameter estimator, implemented as an MLP, takes the plane proxies as input to predict plane parameters. Each plane is represented in polar coordinates \( (r_i, \theta_i, \varphi_i) \) instead of the Cartesian representation, to avoid degeneracies for planes parallel to the axes.

\paragraph{Point Distributor.}
Parameters alone do not define a primitive. Here, we predict the distribution of points on each plane. Leveraging the \( M \) plane parameters generated by the parameter estimator, we use a point distributor head to reconstruct the point distribution on the predicted planes. Specifically, we estimate the polar angles  \((\theta_{ij}, \varphi_{ij})\) of every point, while the radius is derived from these angles and the plane parameters:
\begin{equation}
r_{ij} = \frac{r_i}{\cos(\Delta\varphi) \sin(\theta_{ij}) \sin(\theta_i) + \cos(\theta_{ij}) \cos(\theta_i)},
\end{equation}
where \(\Delta\varphi= \varphi_{ij} - \varphi_i\), and \((r_{ij}, \theta_{ij}, \varphi_{ij})\) represents the polar coordinates of the $j$-th point on the $i$-th plane. 

\paragraph{Primitive Selector.}
As a fixed number of primitives are predicted, not all will comprise the surface. For each predicted plane proxy, the primitive selector assigns a confidence score $\kappa_i \in [0, 1]$, indicating the probability that the primitive is part of the object surface. This score helps distinguish surface-contributing primitives from non-contributing ones. During training, only the selected planar primitives are supervised as detailed in \cref{sec:optimization}.

\subsection{Set Matching and Optimization}
\label{sec:optimization}
\begin{figure}[t]
  \centering
  \includegraphics[width=0.99\linewidth]{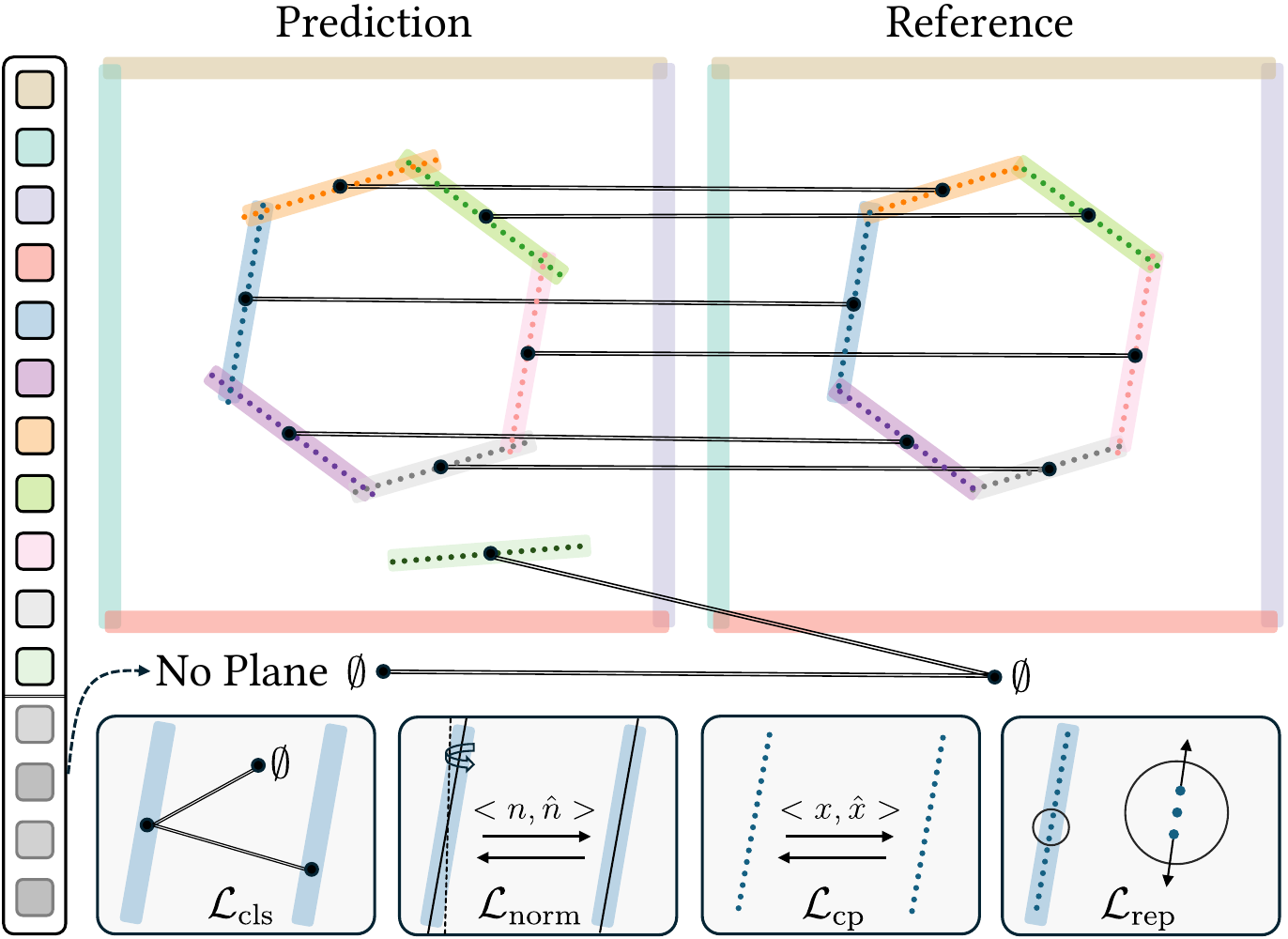}
  \caption{\textbf{Optimization objectives.} Our bipartite matching incorporates four objectives: primitive classification loss ($\mathcal{L}_\mathrm{cls}$) for primitive selection, plane normal loss ($\mathcal{L}_\mathrm{norm}$) for plane parameter estimation, plane chamfer loss ($\mathcal{L}_\mathrm{cp}$) for inlier alignment, and repulsion loss ($\mathcal{L}_\mathrm{rep}$) for uniform point distribution. Some pairs are omitted for brevity.}
  \label{fig:optimization}
\end{figure}

The ground truth consists of a set of unordered planar primitives of varying sizes. Given the predicted primitives, we compare them against the ground truth by constructing a bipartite matching graph between the two sets and applying loss functions to the matched pairs. The network is then optimized using these objectives. \cref{fig:optimization} illustrates the main optimization objectives.

\paragraph{Bipartite Matching.}
We establish correspondences between ground truth planar primitives \( P = \{p_i\}_{i=1}^{M}\) and predicted primitives \(\hat{P} = \{\hat{p}_i\}_{i=1}^{M} \) using bipartite matching~\cite{carion2020end}. The cost matrix for matching is aligned with the loss function to promote stable training. For each matched pair of predicted and ground truth primitives, the matching cost between ground truth primitive \( p_i \) and the prediction at index \( \sigma(i) \) is computed as
\begin{equation}
   C(p_i, \hat{p}_{\sigma(i)})= \underbrace{\mathcal{L}_\mathrm{cls}}_{\text{semantic}} + 
 \underbrace{\beta_1 \mathcal{L}_\mathrm{norm} +\beta_2 \mathcal{L}_\mathrm{cp} + \beta_3 \mathcal{L}_\mathrm{rep}}_{\text{geometric}}.
 \label{eq:cost}  
\end{equation}
The geometric cost incorporates plane normal loss $\mathcal{L}_\mathrm{norm}$, plane completion loss $\mathcal{L}_\mathrm{cp}$, and repulsion loss $\mathcal{L}_\mathrm{rep}$, while the semantic cost is defined by the primitive classification loss $\mathcal{L}_\mathrm{cls}$. The coefficients \( \beta_1 \), \( \beta_2 \), and \( \beta_3 \) weight each respective loss term.

We represent the ground truth planar primitives as a set, padding it with \(\emptyset\) to match the cardinality of the predicted set, where \(\emptyset\) represents the absence of a planar primitive. To establish bipartite matching between these two sets, we search for a permutation \( \sigma \in \Pi \) that minimizes the total matching cost:
\begin{equation}
\hat{\sigma} = \arg \min_{\sigma \in \Pi} \sum_{i}^{M} C(p_i, \hat{p}_{\sigma(i)}).
\end{equation}
This optimal assignment is efficiently computed using the Hungarian algorithm~\cite{kuhn1955hungarian}.

\paragraph{Loss Function.}
\label{para:loss}
Our loss function quantifies discrepancies between prediction and ground truth across three aspects: primitive selection, plane parameter estimation, and point completion. It consists of the following terms:

\textbullet\ \textit{Primitive classification loss} encourages the selection of primitives contributing to the surface while discouraging those not contributing. We define this loss for all predicted primitives as
\begin{equation}
\mathcal{L}_\mathrm{cls} =-\mathbbm{1}_{\{\hat{c}_i = 1\}} \cdot \log \kappa_{\hat{\sigma}(i)}
-\mathbbm{1}_{\{\hat{c}_i = 0\}} \cdot \log (1 - \kappa_{\hat{\sigma}(i)}),
\end{equation}
where \( \kappa_{\hat{\sigma}(i)}\) represents the confidence score predicted by the primitive selector, and $\mathbbm{1}$ is an indicator function that equals 1 if the predicted primitive corresponds to a target primitive and 0 otherwise.

\textbullet\ \textit{Plane normal loss} penalizes the discrepancy between predicted and ground truth plane normals, defined as
\begin{align}
\mathcal{L}_{\mathrm{norm}} = 
& \, \mathbbm{1}_{\{\hat{c}_i = 1\}} \cdot \lambda \left(1 - \cos \left( \angle \left(n_i, \hat{n}_{\hat{\sigma}(i)}\right)\right)\right) + \notag \\
& \, \mathbbm{1}_{\{\hat{c}_i = 1\}} \cdot \| n_i - \hat{n}_{\hat{\sigma}(i)} \|^2,
\end{align}
where \(\ell_2\) distance and cosine similarity are used to measure their similarity, and \(\lambda\) is a scaling factor.

\textbullet\ \textit{Plane chamfer loss} aligns inlier points on each matched primitive pair by assessing the chamfer distance (\textit{CD}):
\begin{equation}
    \mathcal{L}_\mathrm{cp} = \mathbbm{1}_{\{\hat{c}_i = 1\}} \cdot \textit{CD} \left(p_i, \hat{p}_{\sigma(i)}\right).
\end{equation}

\textbullet\ \textit{Overall chamfer loss} further optimizes point distribution. Unlike \( \mathcal{L}_\mathrm{cp} \), which applies to individual planar primitives, this loss term applies to the entire object surface:
\begin{equation}
    \mathcal{L}_\mathrm{co} = \textit{CD} \left(\{ p_i \mid c_i = 1 \}, \{\hat{p}_i \mid \hat{c}_i = 1 \}\right),
\end{equation}
where we aggregate all points from selected primitives and compute their chamfer distance to the ground truth.

\textbullet\ \textit{Repulsion loss} promotes a more uniform point distribution, countering the tendency of points to cluster around the projected center of the plane when trained solely with the plane chamfer loss. The repulsion loss~\cite{yu2018pu} is defined as
\begin{equation}
   \mathcal{L}_\mathrm{rep} = \sum_{i=1}^{T} \sum_{i' \in K(i)} 
   -\left\|\hat{x}_{i'} - \hat{x}_i\right\|\cdot \exp(-\omega \left\|\hat{x}_{i'} - \hat{x}_i\right\|^2), 
\end{equation}
where \(T\) is the number of predicted points on the primitive, \( K(i) \) is the set of indices of the \( k \)-nearest neighbors of point \( \hat{x}_i \), \( \|\hat{x}_i' - \hat{x}_i\| \) represents the distance between generated points \( \hat{x}_i \) and \( \hat{x}_{i'} \), and \(\omega\) is a constant factor.

The final loss is the sum of the above loss terms:
{\small
\begin{equation}
\mathcal{L}_{total}=\sum_{i=1}^{\mathrm{M}}\left(\mathcal{L}_{\mathrm{cls}}^{(i)}+\beta_1\mathcal{L}_{\mathrm{norm}}^{(i)}+\beta_2\mathcal{L}_\mathrm{cp}^{(i)}+\beta_3 \mathcal{L}_{\mathrm{rep}}^{(i)}\right)+\beta_4\mathcal{L}_\mathrm{co},
\label{eq:loss}
\end{equation}
}

\noindent where \( \beta_1 \), \( \beta_2 \), and \( \beta_3 \) are the same as in \cref{eq:cost}, and the additional \( \beta_4 \) weights the overall chamfer loss.

\section{Experiments}
\label{sec:experiments}

\subsection{Experimental Setup}

\paragraph{Dataset.}
To evaluate the performance of PaCo, we conduct experiments on the ABC dataset ~\cite{koch2019abc}, a large-scale collection of CAD models for geometric deep learning. We select 15,339 CAD models with \textit{plane-only} structures (see supplementary material). Following AdaPoinTr~\cite{yu2023adapointr}, we define three occlusion levels (\textit{simple}, \textit{moderate}, and \textit{hard}) with point missing ratios of 25\%, 50\%, and 75\%, respectively. Eight views are generated per model and occlusion level, yielding 368,136 point clouds. We reserve 2,375 CAD models and their associated point clouds for evaluation, ensuring strict separation from the training data. All data are normalized to a unit diagonal length.

\paragraph{Evaluation Metrics.}
\label{sec:metrics}
To assess reconstruction fidelity, we use symmetric chamfer distance \textit{(CD)}, Hausdorff distance \textit{(HD)}, and normal consistency \textit{(NC)}, computed from 10,000 sampled surface points. Additionally, the failure rate \textit{(FR)} is reported as the proportion of unsolvable samples, with these samples evaluated against the unit cube. Unless otherwise specified, models are trained for 200 epochs for ablation studies (\cref{sec:ablation}) and 600 epochs for all other experiments.

\subsection{Evaluation against Completion}

\begin{table*}[htb]
\centering
\caption{\textbf{Quantitative comparison with conventional point cloud completion methods.}  Evaluation is conducted on the reconstructed meshes. \textbf{Bold} numbers indicate the best results, \underline{underlined} the second best. Our method consistently outperforms others across nearly all metrics regardless of the reconstruction solver used. Note: CD, HD, and FR values are scaled by a factor of 100.}
\footnotesize
\begin{tabular}{lccccc|ccccc|cccc}
    \toprule
    \multirow{2}{*}{\raisebox{-0.5ex}{Method}} & \multicolumn{4}{c}{PolyFit \cite{nan2017polyfit}}  & & \multicolumn{4}{c}{COMPOD \cite{sulzer2024concise}}   & & \multicolumn{4}{c}{KSR \cite{bauchet2020kinetic}}  \\
    \cmidrule(lr){2-6} \cmidrule(lr){7-11}\cmidrule(lr){12-15}
    & CD $\downarrow$  & HD $\downarrow$ & NC $\uparrow$ & FR $\downarrow$ & & CD $\downarrow$  & HD $\downarrow$  & NC $\uparrow$ & FR $\downarrow$ & & CD $\downarrow$  & HD $\downarrow$  & NC $\uparrow$  & FR $\downarrow$\\
    \midrule
    PCN \cite{yuan2018pcn} & 14.10 & 20.73 & 0.620  & 71.27 & & 16.27 & 22.20 &  0.562 & 97.87& & 21.61 & 44.23 &0.691 & 11.90\\
    FoldingNet \cite{yang2018foldingnet} & 12.07 & 21.24  &  0.814 & \underline{3.54}&& 7.40 & 18.60&0.805 & \textbf{0.04}& & 18.11  & 42.59 & 0.693 & 1.86\\
    GRNet \cite{xie2020grnet} & 11.98 & 19.84 & 0.769  & 29.61& & 14.19 & 25.72 & 0.738 & 13.15& & 9.18 & 22.01 &0.822 & 10.82\\
    PoinTr \cite{yu2021pointr} & 10.57 & 16.43 & 0.822  & 25.92& & 7.82 & 16.44& 0.774 & 31.34& & 8.14 & 16.33 &0.780& 30.90\\
    AdaPoinTr \cite{yu2023adapointr} & 3.16 & 7.36 & 0.920 & 5.89 & & 3.25& 8.84  &0.921 & 1.32& & 3.24 & 8.86 &0.927 & \underline{0.27}\\
    ODGNet \cite{cai2024orthogonal}  & \underline{2.73} & \underline{6.41} & \underline{0.933}  & 4.28& & \underline{3.22} & \underline{8.01} &\underline{0.927}& 1.05& & \underline{2.90} & \underline{8.56} &\underline{0.934} & 0.36\\
    PaCo (ours) & \textbf{1.87} & \textbf{4.09} & \textbf{0.943}  & \textbf{0.48} & & \textbf{1.94} & \textbf{4.42} & \textbf{0.940} & \underline{0.25} & & \textbf{1.91} & \textbf{4.14} & \textbf{0.940} & \textbf{0.25} \\
    \bottomrule
\end{tabular}
\label{tab:completion}
\end{table*}

Since the completion is dedicated to surface reconstruction, we benchmark our completion scheme against state-of-the-art point cloud completion methods by feeding the completed points to polygonal surface reconstruction solvers and evaluating the reconstructed surfaces. In this evaluation, we include six point cloud completion methods, PCN~\cite{yuan2018pcn}, FoldingNet~\cite{yang2018foldingnet}, GRNet~\cite{xie2020grnet}, PoinTr~\cite{yu2021pointr}, AdaPoinTr~\cite{yu2023adapointr}, and ODGNet~\cite{cai2024orthogonal}, along with three polygonal surface reconstruction solvers, PolyFit~\cite{nan2017polyfit}, KSR~\cite{bauchet2020kinetic}, and COMPOD~\cite{sulzer2024concise}. Unlike the competing methods that lack explicit structures in their predictions, PaCo generates parametric planar primitives that can be \textit{directly} used by the solvers, while GoCoPP~\cite{yu2022finding} is used to extract planar primitives for the other completion methods.

\begin{figure*}
  \centering
  \includegraphics[width=0.95\linewidth]{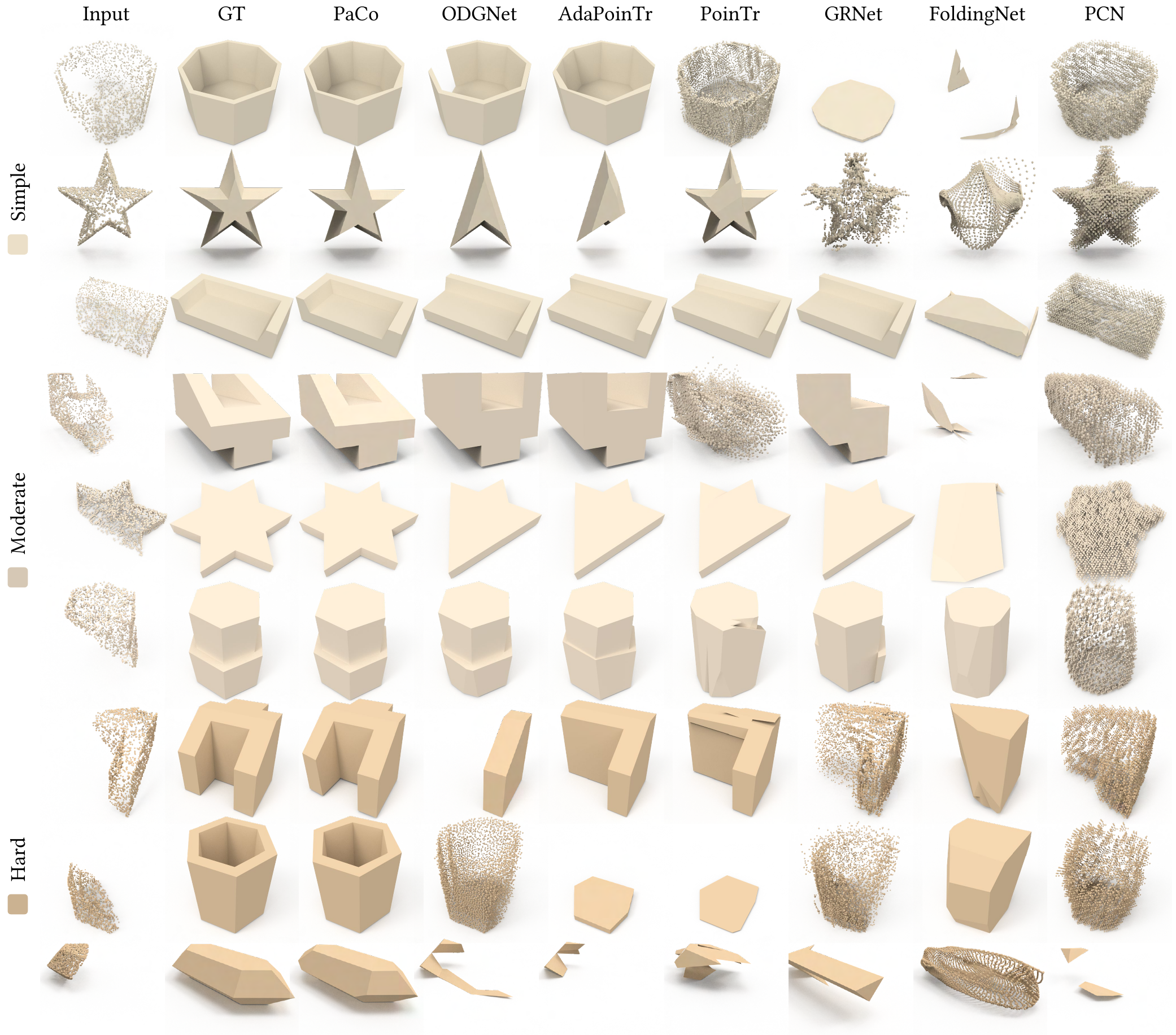}
  \caption{\textbf{Qualitative comparison with conventional point cloud completion methods.} Simple, moderate, and hard denote missing ratios of 25\%, 50\%, and 75\%, respectively. For cases where reconstruction fails, the completed points are shown instead. Our method excels in recovering geometric structures with planar primitives directly usable by the reconstruction solver.}
  \label{fig:completion}
\end{figure*}

\cref{tab:completion} presents the quantitative results. Our method outperforms others across \textit{all} fidelity metrics by substantial margins and also achieves the \textit{lowest} failure rate, regardless of the reconstruction solver used. \cref{fig:completion} provides qualitative results with the PolyFit solver: our method produces surface geometries closest to the ground truth, whereas other completion methods frequently miss structures or even completely fail. PaCo's performance is also the \textit{most consistent} across different occlusion levels.

\subsection{Evaluation against Reconstruction}
\label{sec:reconstruction}

In addition to the completion-and-reconstruction comparison, we also directly compare our approach with both traditional and recent neural reconstruction methods without the point cloud completion step. This evaluation helps to understand the role of completion in achieving faithful surface reconstruction. The traditional methods involved in this analysis include PolyFit~\cite{nan2017polyfit}, KSR~\cite{bauchet2020kinetic}, and COMPOD~\cite{sulzer2024concise}, while the neural methods include BSP-Net~\cite{chen2020bspnet} and SECAD-Net~\cite{li2023secad}. Since both neural methods require voxel input, we prepare voxels with the same level of incompleteness for consistency. For our approach, we choose the PolyFit solver as it shows strong performance in \cref{tab:completion}.

\begin{figure}
  \centering
  \includegraphics[width=0.99\linewidth]{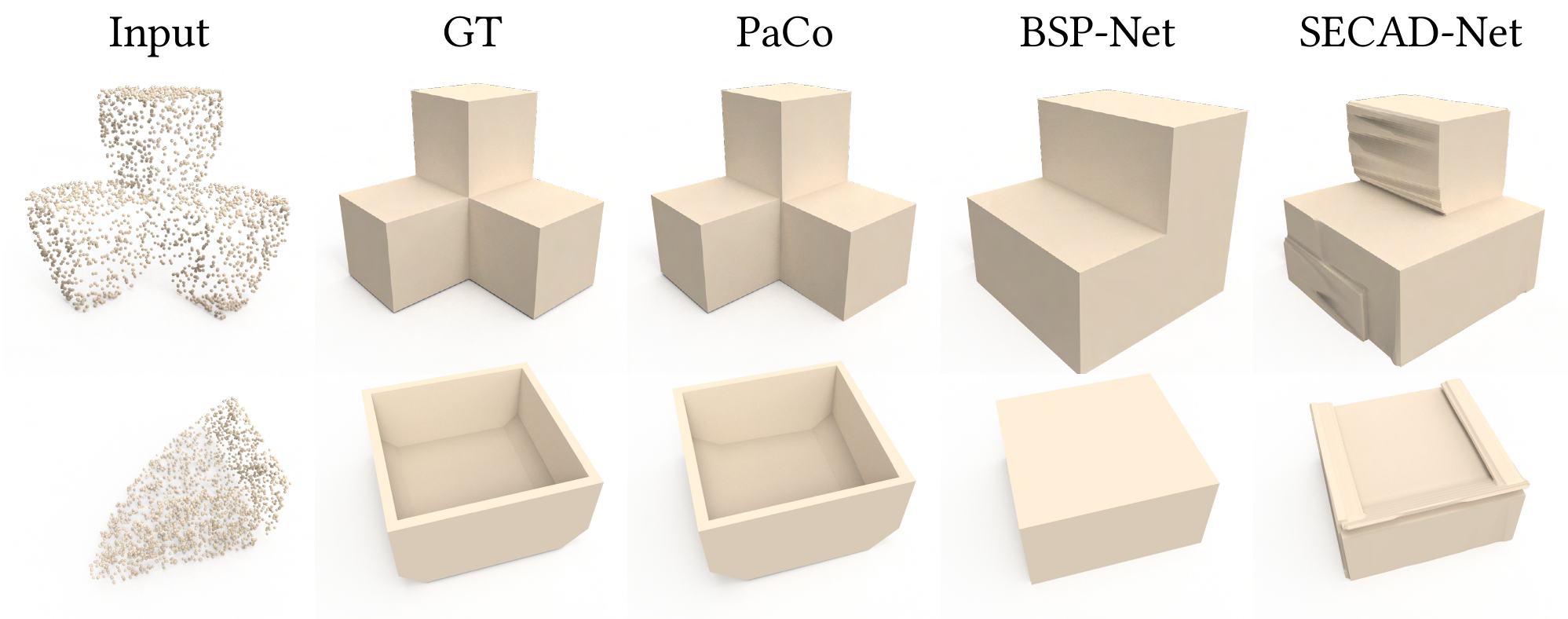}
  \caption{\textbf{Qualitative comparison with neural reconstruction methods directly from incomplete inputs}. Our method demonstrates stronger capability in terms of structure recovery.}
  \label{fig:reconstruction}
\end{figure}

\cref{tab:reconstruction} presents the quantitative comparison. All traditional reconstruction methods fail on a significant portion of inputs, hindered by the missing structures in the input points. While the two neural reconstruction methods manage to produce some geometry, they still fail to capture the main structures of the object surfaces. \cref{fig:reconstruction} shows qualitative examples, where BSP-Net tends to produce over-simplified meshes, and both competing methods struggle to generalize. In contrast, our method demonstrates the \textit{strongest} capability in structure recovery. This comparison also reveals that direct reconstruction of polygonal surfaces from incomplete data is less effective than the completion-and-reconstruction strategy, as evidenced by the performance gap between \cref{tab:completion} and \cref{tab:reconstruction}.

\begin{table}
  \centering
  \caption{\textbf{Quantitative comparison with reconstruction methods.} Neural approaches are in gray. Traditional methods fail on a significant portion of inputs, while the competing neural methods struggle to preserve structural integrity, both underperform our completion-and-reconstruction strategy.}
  \footnotesize
  \begin{tabular}{lcccc}
    \toprule
    Method & {CD $\downarrow$} & {HD $\downarrow$} & {NC$\uparrow$} & {FR $\downarrow$} \\
    \midrule
    PolyFit \cite{nan2017polyfit} & 15.11 & 24.26 & 0.618 & 63.28 \\
    KSR\cite{bauchet2020kinetic} & 9.18 & 22.01 & 0.822 & 10.82 \\
    COMPOD \cite{sulzer2024concise} & 6.87 & 18.88 & 0.826 & 9.64 \\
    \midrule
    \rowcolor{lightgray} BSP-Net \cite{chen2020bspnet} & 7.69 & 20.14 & 0.652 & \textbf{0.00} \\
    \rowcolor{lightgray} SECAD-Net \cite{li2023secad} & \underline{4.35} & \underline{11.33} & \underline{0.850} & \textbf{0.00} \\
    \rowcolor{lightgray} PaCo (ours) & \textbf{1.87} & \textbf{4.09}  & \textbf{0.943} & \underline{0.48} \\
    \bottomrule
  \end{tabular}
  \label{tab:reconstruction}
\end{table}

\subsection{Evaluation against Simplification}

We further compare our approach with two geometric simplification methods: the classic QEM~\cite{garland1997qem} and the recent RoLoPM~\cite{chen2023robust}. 
We use PolyFit as the reconstruction solver for our approach, and ODGNet~\cite{cai2024orthogonal} to complete the points for Poisson reconstruction~\cite{kazhdan2013screened}, given their strong performance in \cref{tab:completion}. Simplification is then applied to the Poisson results. In addition to mesh fidelity, we report face count ($\#f$) and vertex count ($\#v$) as indicators of geometric complexity. To ensure a fair comparison, we set the target face counts for the simplification methods to closely match our triangulated results.

\begin{figure}
  \centering
  \includegraphics[width=0.99\linewidth]{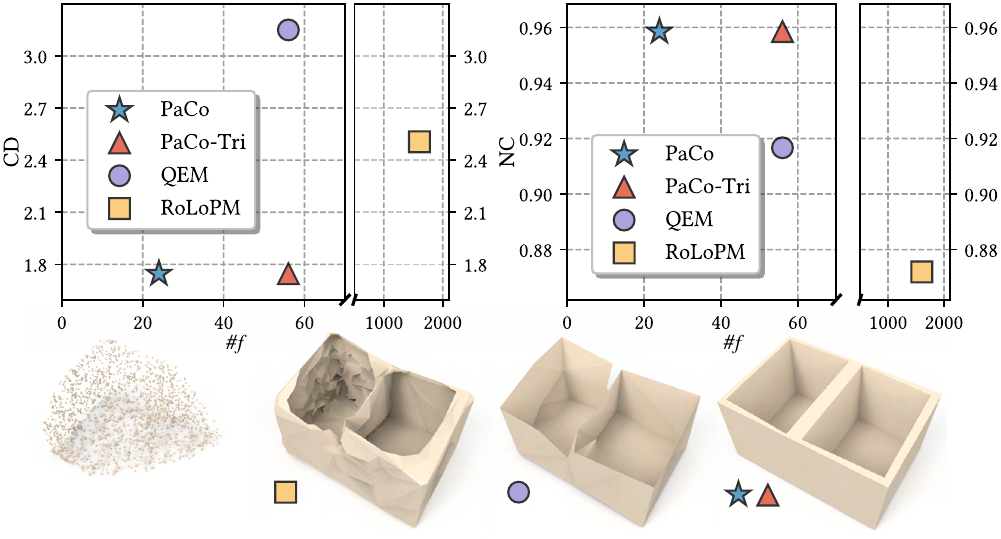}
  \caption{\textbf{Qualitative comparison with simplification methods.} Simplifications are applied to the Poisson reconstruction from the completed points. PaCo-Tri denotes ours triangulated. Our result demonstrates greater regularity and has the fewest faces.}
  \label{fig:simplification}
\end{figure}

\cref{tab:simplification} presents the quantitative results on 50 randomly selected samples, where our approach achieves the \textit{highest} performance while using the \textit{fewest} faces. \cref{fig:simplification} provides a qualitative comparison with the two simplification methods, showing that our result has significantly greater regularity.

\begin{table}[t]
\centering
\caption{\textbf{Quantitative comparison with simplification methods}. $\#f$ and $\#v$ denote the number of faces and vertices, respectively. All metrics are measured on 50 randomly selected samples. ``39 tri.'' denotes the average triangle number after triangulation.}
\footnotesize
\begin{tabular}{lcc|ccc}
\toprule
Method & $\#f$ & $\#v$ & CD $\downarrow$ & HD $\downarrow$ & NC $\uparrow$ \\
\midrule
QEM~\cite{garland1997qem}      & 39 & 22 & 2.67 & 11.32 & 0.900 \\
RoLoPM~\cite{chen2023robust}   & 44 & 23 & 2.43 & 9.05  & 0.918 \\
PaCo (ours)                    & \textbf{13} (39 tri.) & 22 & \textbf{1.72} & \textbf{4.13} & \textbf{0.934} \\
\bottomrule
\end{tabular}
\label{tab:simplification}
\end{table}

\subsection{Robustness Analysis}
\label{sec:robustness}
\paragraph{Noise.} 

\begin{figure}[t]
  \centering
  \includegraphics[width=0.99\linewidth]{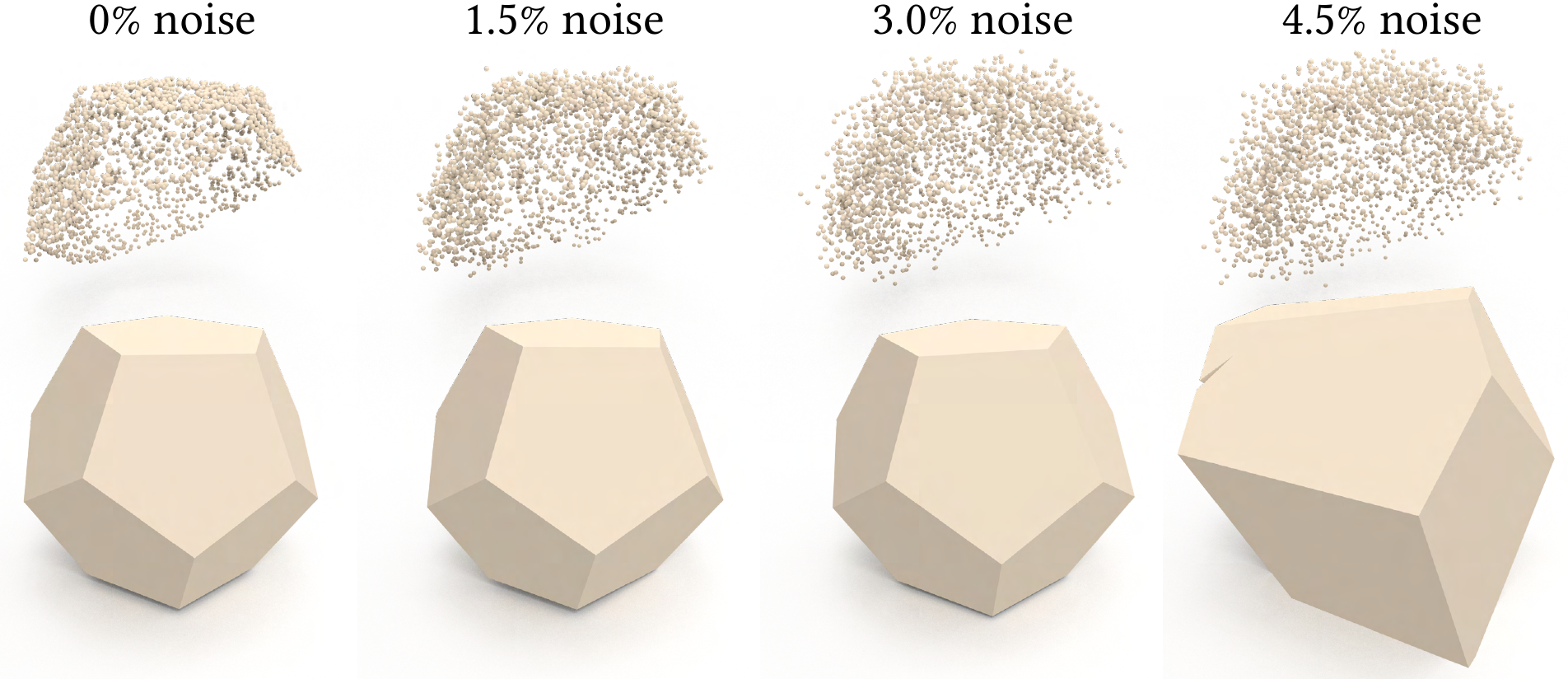}
  \caption{\textbf{Qualitative example on robustness against noise.} Reliable reconstruction is obtained, up to a prohibitive noise level of 4.5\% Gaussian noise where structural changes become evident.}
  \label{fig:noise}
\end{figure}

\cref{tab:noise} presents quantitative results of our method under different noise levels, with \cref{fig:noise} showing a series of qualitative examples. Note that models are trained at respective noise levels for 200 epochs, with initial planar structures intact. Gaussian noise at 1.5\% of the object’s diagonal length results in a slight increase in errors, while raising the noise to 3.0\% does not significantly degrade performance. The failure rate remains stable as noise increases, up to a prohibitive level of 4.5\% where structural changes become evident. These results reinforce our focus on parametric recovery over completing individual points.

\begin{table}[t]
  \centering
  \caption{\textbf{Quantitative evaluation on robustness against noise.} Ratios denote Gaussian noise defined relative to the object's diagonal length. Models are trained on respective noise levels.}
  \footnotesize
  \begin{tabular}{lccccc}
    \toprule
    Noise & CD $\downarrow$ & HD $\downarrow$ & NC $\uparrow$ & FR $\downarrow$ &  \\
    \midrule
     4.5\% &3.75 & 7.97 &  0.872 &  0.65\\
     3.0\% &2.32 & \textbf{5.06} &  0.925 &  0.48\\
     1.5\% &2.29 & 5.07 & 0.925  &  \textbf{0.44}\\
     0.0\% &\textbf{2.11} & 5.26 &  \textbf{0.941} &  0.48\\
    \bottomrule
  \end{tabular}
  \label{tab:noise}
\end{table}

\paragraph{Incompleteness.}

\cref{tab:incompleteness} quantitatively compares our method with others across different levels of incompleteness. At the \textit{simple} incompleteness level, the performance gain of our method is relatively modest. However, on more challenging data featuring larger missing areas (\textit{moderate} and \textit{hard}), all competing methods show a steep decline in performance, while ours remains largely stable. This robustness results in the pronounced performance advantage on the \textit{hard} samples and superior overall performance. \cref{fig:completion} visually demonstrates this trend: with greater incompleteness, our method continues to produce plausible reconstructions, whereas other methods struggle to preserve structural integrity as they primarily focus on recovering individual points. In the supplementary material, we show that PaCo also effectively processes real-scanned airborne LiDAR data for building reconstruction.

\begin{table}
    \centering
    \caption{\textbf{Performance comparison in terms of data incompleteness.} Our method performs competitively on simple data (S, -25\%), and excels on moderate (M, -50\%) and hard (H, -75\%) data.}
    \footnotesize
    \resizebox{\linewidth}{!}{
    \begin{tabular}{lccc|ccc}
        \toprule
        \multirow{2}{*}{\raisebox{-0.5ex}{Method}} & \multicolumn{3}{c|}{CD $\downarrow$} & \multicolumn{3}{c}{NC $\uparrow$} \\
        \cmidrule(lr){2-4} \cmidrule(lr){5-7}
        & S & M & H & S & M & H \\
        \midrule
        PCN \cite{yuan2018pcn} & 13.06 & 13.59 & 15.65 & 0.654 & 0.631 & 0.577 \\
        FoldingNet \cite{yang2018foldingnet} & 11.22 & 11.67 & 13.33 & 0.840 & 0.825 & 0.776 \\
        GRNet \cite{xie2020grnet} & 10.23 & 11.76 & 13.97 & 0.849 & 0.780 & 0.679 \\
        PoinTr \cite{yu2021pointr} & 10.05 & 10.39 & 11.28 & 0.846 & 0.830 & 0.789 \\
        AdaPoinTr \cite{yu2023adapointr} & 1.55 & 2.12 & 5.81 & \underline{0.965} & 0.949 & 0.846 \\
        ODGNet \cite{cai2024orthogonal} & \textbf{1.48} & \underline{2.01} & \underline{4.72} & \textbf{0.969} & \underline{0.950} & \underline{0.879} \\
        PaCo (ours) & \underline{1.52} & \textbf{1.66} & \textbf{2.43} & 0.959 & \textbf{0.952} & \textbf{0.919} \\
        \bottomrule
    \end{tabular}
  }
  \label{tab:incompleteness}
\end{table}

\subsection{Ablation Studies}
\label{sec:ablation}

\begin{figure}
  \centering
  \includegraphics[width=0.99\linewidth]{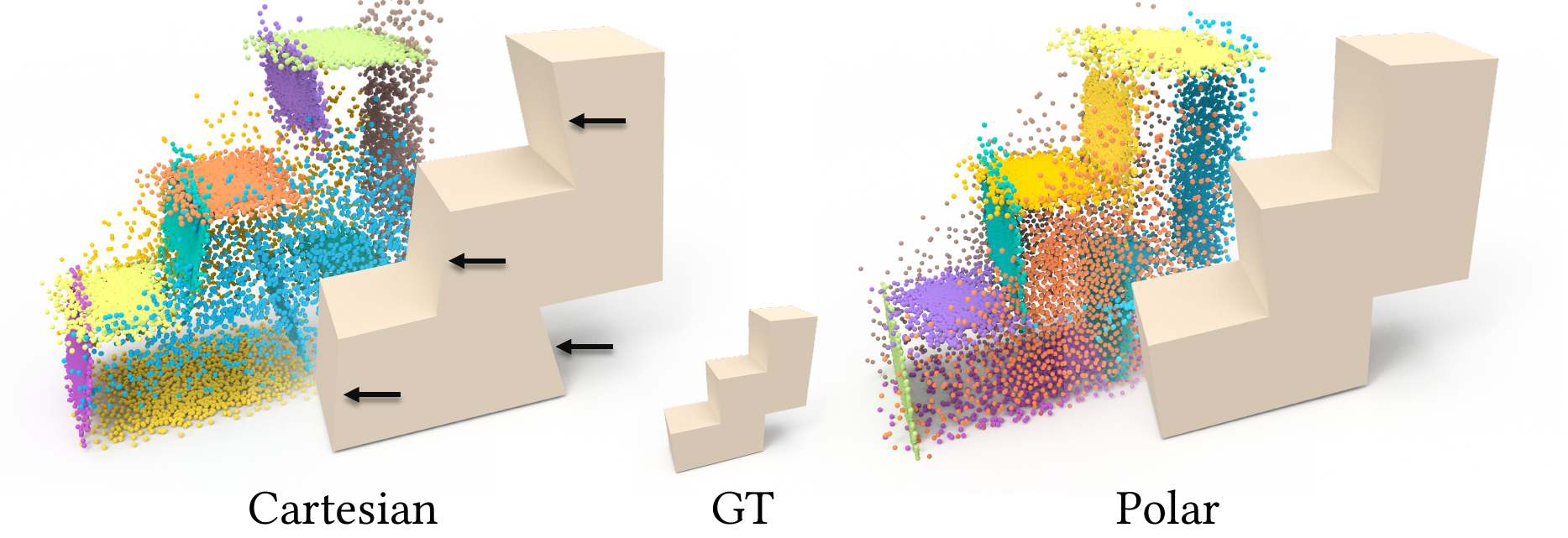}
  \caption{\textbf{Impact of coordinate representation.} Our polar representation facilitates the formulation of axis-aligned planar primitives, producing structures of greater regularity than that of the Cartesian counterpart.}
  \label{fig:coordinate}
\end{figure}

\paragraph{Network Architecture.}
\cref{tab:ablation_component} presents the contributions of key components in our network architecture. Replacing the polar representation with the commonly used Cartesian representation results in a substantial performance drop. \cref{fig:coordinate} visually compares these representations, showing that the polar form enables the formation of primitives perpendicular to the ground, producing structures with greater regularity. In contrast, the Cartesian representation struggles to achieve this structural consistency. Similarly, removing plane proxies in the encoding hierarchy hinders performance, as point proxies alone cannot provide structured feature aggregation as needed. Lastly, substituting sum with mean pooling in the aggregation step leads to the loss of scale information across proxies, resulting in a marked performance reduction.

\begin{table}
  \centering
  \caption{\textbf{Ablation of network components.} 
  Our method is optimally configured with polar coordinate representation, plane proxies, and sum pooling aggregation. ``-'' denotes not applicable.}
  \footnotesize
  \begin{tabular}{ccc|ccc}
    \toprule
    \multicolumn{3}{c|}{Component} & \multirow{2}{*}{\raisebox{-0.5ex}{CD $\downarrow$}} & \multirow{2}{*}{\raisebox{-0.5ex}{HD $\downarrow$}} & \multirow{2}{*}{\raisebox{-0.5ex}{NC $\uparrow$}} \\
    \cmidrule(lr){1-3}
    Polar & Plane Proxy & Aggregation & & & \\
    \midrule
      & \checkmark & sum & 3.34 & 7.87 &  0.879\\
    \checkmark &  & - & 2.45 & 6.49 & 0.916\\
    \checkmark & \checkmark & mean & 2.31 & 5.90 & 0.924\\
    \checkmark & \checkmark & sum & \textbf{2.11} & \textbf{5.26} & \textbf{0.941}\\
    \bottomrule
  \end{tabular}
  \label{tab:ablation_component}
\end{table}

\paragraph{Loss Terms.}

\cref{tab:ablation_loss} further presents the contributions of the optional loss terms $\mathcal{L}_\mathrm{rep}$, $\mathcal{L}_\mathrm{co}$, and $\mathcal{L}_\mathrm{norm}$ in \cref{eq:loss}. Removing any of these terms leads to a significant and consistent performance drop across all metrics. This reveals the importance of maintaining a repulsive and reference-aligned point distribution and emphasizes the critical role of accurate plane parameter estimation.

\begin{table}
  \centering
  \caption{\textbf{Ablation of loss terms.} Our method performs the best when the repulsion loss ($\mathcal{L}_\mathrm{rep}$), overall chamfer loss ($\mathcal{L}_\mathrm{co}$), and plane normal loss ($\mathcal{L}_\mathrm{norm}$) are all present.}
  \footnotesize
  \begin{tabular}{ccc|ccc}
    \toprule
    \multicolumn{3}{c|}{Term} & \multirow{2}{*}{\raisebox{-0.5ex}{CD $\downarrow$}} & \multirow{2}{*}{\raisebox{-0.5ex}{HD $\downarrow$}} & \multirow{2}{*}{\raisebox{-0.5ex}{NC $\uparrow$}} \\
    \cmidrule(lr){1-3}
    $\mathcal{L}_\mathrm{rep}$ & $\mathcal{L}_\mathrm{co}$ & $\mathcal{L}_\mathrm{norm}$ & & & \\
    \midrule
      & \checkmark & \checkmark & 2.48 & 6.37 & 0.916 \\
    \checkmark &  & \checkmark & 2.45 & 6.15 & 0.918 \\
    \checkmark & \checkmark & & 2.22 & 5.42 & 0.928 \\
    \checkmark & \checkmark & \checkmark & \textbf{2.11} & \textbf{5.26} & \textbf{0.941} \\
    \bottomrule
  \end{tabular}
  \label{tab:ablation_loss}
\end{table}
\section{Conclusion}
\label{sec:conclusion}

We have introduced parametric completion, a new paradigm for point cloud completion, and presented PaCo, a novel framework that leverages this paradigm to address the challenges of polygonal surface reconstruction. Unlike conventional methods that prioritize individual point recovery, PaCo recovers structured, parametric planar primitives, enabling accurate and robust reconstruction of polygonal surfaces from incomplete point cloud data. This paradigm shift from point-based to parametric representation bridges the gap between shape completion and polygonal surface reconstruction and establishes a new standard for recovering structured geometries from incomplete data.

\paragraph{Limitations.}
PaCo is designed specifically for polygonal surface reconstruction and is not optimized for conventional point cloud completion tasks that prioritize the distribution of individual points. Its hierarchical encoding is based on the presence of planar structures in the input, which can limit the ability of PaCo to form meaningful proxies if such structures are absent. However, we did not encounter such failures in our experiments, even in challenging scenarios. Besides, the current implementation of PaCo is restricted to point clouds comprising only planar primitives, and it may fail for objects with non-planar geometries.

\paragraph{Future Work.}
A natural extension of PaCo would be to incorporate additional parametric shape primitives, such as spheres and cylinders. We anticipate that this extension will further strengthen the connection between point cloud completion and surface reconstruction, advancing toward a unified framework for recovering complex object geometries from incomplete data.
\par
\section*{Acknowledgment}
This work was supported by TUM Georg Nemetschek Institute under the AI4TWINNING project. We thank anonymous reviewers for their constructive comments.

{
    \small
    \bibliographystyle{ieeenat_fullname}
    \bibliography{main}

\begin{thebibliography}{50}
\providecommand{\natexlab}[1]{#1}
\providecommand{\url}[1]{\texttt{#1}}
\expandafter\ifx\csname urlstyle\endcsname\relax
  \providecommand{\doi}[1]{doi: #1}\else
  \providecommand{\doi}{doi: \begingroup \urlstyle{rm}\Url}\fi

\bibitem[Bauchet and Lafarge(2020)]{bauchet2020kinetic}
Jean-Philippe Bauchet and Florent Lafarge.
\newblock Kinetic shape reconstruction.
\newblock \emph{ACM TOG}, 39\penalty0 (5):\penalty0 1--14, 2020.

\bibitem[Botsch et~al.(2010)Botsch, Kobbelt, Pauly, Alliez, and L{\'e}vy]{botsch2010polygon}
Mario Botsch, Leif Kobbelt, Mark Pauly, Pierre Alliez, and Bruno L{\'e}vy.
\newblock \emph{Polygon mesh processing}.
\newblock AK Peters/CRC Press, Natick, MA, USA, 2010.

\bibitem[Boulch and Marlet(2022)]{boulch2022poco}
Alexandre Boulch and Renaud Marlet.
\newblock {POCO}: Point convolution for surface reconstruction.
\newblock In \emph{CVPR}, pages 6302--6314, 2022.

\bibitem[Cai et~al.(2024)Cai, Scott, Li, and Wang]{cai2024orthogonal}
Pingping Cai, Deja Scott, Xiaoguang Li, and Song Wang.
\newblock Orthogonal dictionary guided shape completion network for point cloud.
\newblock In \emph{AAAI}, pages 864--872, 2024.

\bibitem[Carion et~al.(2020)Carion, Massa, Synnaeve, Usunier, Kirillov, and Zagoruyko]{carion2020end}
Nicolas Carion, Francisco Massa, Gabriel Synnaeve, Nicolas Usunier, Alexander Kirillov, and Sergey Zagoruyko.
\newblock End-to-end object detection with transformers.
\newblock In \emph{ECCV}, pages 213--229, 2020.

\bibitem[Chen et~al.(2020)Chen, Tagliasacchi, and Zhang]{chen2020bspnet}
Zhiqin Chen, Andrea Tagliasacchi, and Hao Zhang.
\newblock {BSP-Net: Generating compact meshes via binary space partitioning}.
\newblock In \emph{CVPR}, pages 45--54, 2020.

\bibitem[Chen et~al.(2022)Chen, Ledoux, Khademi, and Nan]{chen2022points2poly}
Zhaiyu Chen, Hugo Ledoux, Seyran Khademi, and Liangliang Nan.
\newblock Reconstructing compact building models from point clouds using deep implicit fields.
\newblock \emph{ISPRS Journal of Photogrammetry and Remote Sensing}, 194:\penalty0 58--73, 2022.

\bibitem[Chen et~al.(2023)Chen, Pan, Wu, Vouga, and Gao]{chen2023robust}
Zhen Chen, Zherong Pan, Kui Wu, Etienne Vouga, and Xifeng Gao.
\newblock Robust low-poly meshing for general {3D} models.
\newblock \emph{ACM TOG}, 42\penalty0 (4):\penalty0 1--20, 2023.

\bibitem[Chen et~al.(2024)Chen, Shi, Nan, Xiong, and Zhu]{chen2024polygnn}
Zhaiyu Chen, Yilei Shi, Liangliang Nan, Zhitong Xiong, and Xiao~Xiang Zhu.
\newblock {PolyGNN}: Polyhedron-based graph neural network for {3D} building reconstruction from point clouds.
\newblock \emph{ISPRS Journal of Photogrammetry and Remote Sensing}, 218:\penalty0 693--706, 2024.

\bibitem[Cohen-Steiner et~al.(2004)Cohen-Steiner, Alliez, and Desbrun]{cohen2004vsa}
David Cohen-Steiner, Pierre Alliez, and Mathieu Desbrun.
\newblock Variational shape approximation.
\newblock In \emph{ACM SIGGRAPH 2004 Papers}, page 905–914, New York, NY, USA, 2004. Association for Computing Machinery.

\bibitem[Dai et~al.(2017)Dai, Qi, and Nie{\ss}ner]{dai2017}
Angela Dai, Charles~R. Qi, and Matthias Nie{\ss}ner.
\newblock {Shape completion using 3D encoder-predictor CNNs and shape synthesis}.
\newblock In \emph{CVPR}, pages 5868--5877, 2017.

\bibitem[Erler et~al.(2020)Erler, Guerrero, Ohrhallinger, Mitra, and Wimmer]{erler2020points2surf}
Philipp Erler, Paul Guerrero, Stefan Ohrhallinger, Niloy~J Mitra, and Michael Wimmer.
\newblock {Points2Surf}: Learning implicit surfaces from point clouds.
\newblock In \emph{ECCV}, pages 108--124, 2020.

\bibitem[Gao et~al.(2022)Gao, Wu, and Pan]{gao2022lowpoly}
Xifeng Gao, Kui Wu, and Zherong Pan.
\newblock Low-poly mesh generation for building models.
\newblock In \emph{ACM SIGGRAPH 2022 Conference Proceedings}, New York, NY, USA, 2022. Association for Computing Machinery.

\bibitem[Garland and Heckbert(1997)]{garland1997qem}
Michael Garland and Paul~S Heckbert.
\newblock Surface simplification using quadric error metrics.
\newblock In \emph{Proceedings of the 24th annual conference on Computer graphics and interactive techniques}, pages 209--216, 1997.

\bibitem[Huang et~al.(2022{\natexlab{a}})Huang, Chen, and Hu]{huang2022neural}
Jiahui Huang, Hao-Xiang Chen, and Shi-Min Hu.
\newblock A neural {Galerkin} solver for accurate surface reconstruction.
\newblock \emph{ACM TOG}, 41\penalty0 (6):\penalty0 1--16, 2022{\natexlab{a}}.

\bibitem[Huang et~al.(2022{\natexlab{b}})Huang, Stoter, Peters, and Nan]{huang2022city3d}
Jin Huang, Jantien Stoter, Ravi Peters, and Liangliang Nan.
\newblock {City3D}: Large-scale building reconstruction from airborne {LiDAR} point clouds.
\newblock \emph{Remote Sensing}, 14\penalty0 (9), 2022{\natexlab{b}}.

\bibitem[Huang et~al.(2023)Huang, Gojcic, Atzmon, Litany, Fidler, and Williams]{huang2023neural}
Jiahui Huang, Zan Gojcic, Matan Atzmon, Or Litany, Sanja Fidler, and Francis Williams.
\newblock Neural kernel surface reconstruction.
\newblock In \emph{CVPR}, pages 4369--4379, 2023.

\bibitem[Huang et~al.(2024)Huang, Wen, Wang, Ren, and Jia]{huang2024surface}
Zhangjin Huang, Yuxin Wen, Zihao Wang, Jinjuan Ren, and Kui Jia.
\newblock Surface reconstruction from point clouds: A survey and a benchmark.
\newblock \emph{IEEE TPAMI}, 2024.

\bibitem[Kazhdan and Hoppe(2013)]{kazhdan2013screened}
Michael Kazhdan and Hugues Hoppe.
\newblock Screened {Poisson} surface reconstruction.
\newblock \emph{ACM TOG}, 32\penalty0 (3):\penalty0 1--13, 2013.

\bibitem[Koch et~al.(2019)Koch, Matveev, Jiang, Williams, Artemov, Burnaev, Alexa, Zorin, and Panozzo]{koch2019abc}
Sebastian Koch, Albert Matveev, Zhongshi Jiang, Francis Williams, Alexey Artemov, Evgeny Burnaev, Marc Alexa, Denis Zorin, and Daniele Panozzo.
\newblock {ABC}: A big {CAD} model dataset for geometric deep learning.
\newblock In \emph{CVPR}, pages 9601--9611, 2019.

\bibitem[Kuhn(1955)]{kuhn1955hungarian}
Harold~W Kuhn.
\newblock The {Hungarian} method for the assignment problem.
\newblock \emph{Naval research logistics quarterly}, 2\penalty0 (1-2):\penalty0 83--97, 1955.

\bibitem[Li and Nan(2021)]{li2021simplification}
Minglei Li and Liangliang Nan.
\newblock Feature-preserving {3D} mesh simplification for urban buildings.
\newblock \emph{ISPRS Journal of Photogrammetry and Remote Sensing}, 173:\penalty0 135--150, 2021.

\bibitem[Li et~al.(2023{\natexlab{a}})Li, Guo, Zhang, and Yan]{li2023secad}
Pu Li, Jianwei Guo, Xiaopeng Zhang, and Dong-Ming Yan.
\newblock {SECAD-Net}: Self-supervised {CAD} reconstruction by learning sketch-extrude operations.
\newblock In \emph{CVPR}, pages 16816--16826, 2023{\natexlab{a}}.

\bibitem[Li et~al.(2023{\natexlab{b}})Li, Gao, Tan, and Wei]{li2023proxyformer}
Shanshan Li, Pan Gao, Xiaoyang Tan, and Mingqiang Wei.
\newblock {ProxyFormer}: Proxy alignment assisted point cloud completion with missing part sensitive transformer.
\newblock In \emph{CVPR}, pages 9466--9475, 2023{\natexlab{b}}.

\bibitem[Liu et~al.(2024)Liu, Obukhov, Wegner, and Schindler]{liu2024point2cad}
Yujia Liu, Anton Obukhov, Jan~Dirk Wegner, and Konrad Schindler.
\newblock {Point2CAD}: Reverse engineering {CAD} models from {3D} point clouds.
\newblock In \emph{CVPR}, pages 3763--3772, 2024.

\bibitem[Nan and Wonka(2017)]{nan2017polyfit}
Liangliang Nan and Peter Wonka.
\newblock {PolyFit: Polygonal surface reconstruction from point clouds}.
\newblock In \emph{ICCV}, pages 2353--2361, 2017.

\bibitem[Qi et~al.(2017{\natexlab{a}})Qi, Su, Mo, and Guibas]{qi2017pointnet}
Charles~R. Qi, Hao Su, Kaichun Mo, and Leonidas~J. Guibas.
\newblock {PointNet: Deep learning on point sets for 3D classification and segmentation}.
\newblock In \emph{CVPR}, pages 652--660, 2017{\natexlab{a}}.

\bibitem[Qi et~al.(2017{\natexlab{b}})Qi, Yi, Su, and Guibas]{qi2017pointnet++}
Charles~R. Qi, Li Yi, Hao Su, and Leonidas~J. Guibas.
\newblock {PointNet++: Deep hierarchical feature learning on point sets in a metric space}.
\newblock In \emph{NeurIPS}, 2017{\natexlab{b}}.

\bibitem[Ren et~al.(2021)Ren, Zheng, Cai, Li, Jiang, Cai, Zhang, Pan, Zhang, Zhao, et~al.]{ren2021csg}
Daxuan Ren, Jianmin Zheng, Jianfei Cai, Jiatong Li, Haiyong Jiang, Zhongang Cai, Junzhe Zhang, Liang Pan, Mingyuan Zhang, Haiyu Zhao, et~al.
\newblock {CSG-Stump}: A learning friendly {CSG}-like representation for interpretable shape parsing.
\newblock In \emph{ICCV}, pages 12478--12487, 2021.

\bibitem[Salinas et~al.(2015)Salinas, Lafarge, and Alliez]{salinas2015samd}
D. Salinas, F. Lafarge, and P. Alliez.
\newblock Structure-aware mesh decimation.
\newblock \emph{Comput. Graph. Forum}, 34\penalty0 (6):\penalty0 211--227, 2015.

\bibitem[Schnabel et~al.(2007)Schnabel, Wahl, and Klein]{schnabel2007efficient}
Ruwen Schnabel, Roland Wahl, and Reinhard Klein.
\newblock {Efficient RANSAC for point-cloud shape detection}.
\newblock \emph{Comput. Graph. Forum}, 26\penalty0 (2):\penalty0 214--226, 2007.

\bibitem[Sulzer and Lafarge(2024)]{sulzer2024concise}
Raphael Sulzer and Florent Lafarge.
\newblock Concise plane arrangements for low-poly surface and volume modelling.
\newblock \emph{ECCV}, 2024.

\bibitem[Tang et~al.(2022)Tang, Gong, Yi, Xie, and Ma]{tang2022lake}
Junshu Tang, Zhijun Gong, Ran Yi, Yuan Xie, and Lizhuang Ma.
\newblock {LAKe-Net}: Topology-aware point cloud completion by localizing aligned keypoints.
\newblock In \emph{CVPR}, pages 1726--1735, 2022.

\bibitem[Tesema et~al.(2023)Tesema, Hill, Jones, Ahmad, and Tam]{tesema2023point}
Keneni~W Tesema, Lyndon Hill, Mark~W Jones, Muneeb~I Ahmad, and Gary~KL Tam.
\newblock Point cloud completion: A survey.
\newblock \emph{IEEE TVCG}, 2023.

\bibitem[Thomas et~al.(2019)Thomas, Qi, Deschaud, Marcotegui, Goulette, and Guibas]{thomas2019kpconv}
Hugues Thomas, Charles~R. Qi, Jean-Emmanuel Deschaud, Beatriz Marcotegui, François Goulette, and Leonidas~J. Guibas.
\newblock {KPConv: Flexible and deformable convolution for point clouds}.
\newblock In \emph{ICCV}, pages 6411--6420, 2019.

\bibitem[Vaswani et~al.(2017)Vaswani, Shazeer, Parmar, Uszkoreit, Jones, Gomez, Kaiser, and Polosukhin]{vaswani2017attention}
Ashish Vaswani, Noam Shazeer, Niki Parmar, Jakob Uszkoreit, Llion Jones, Aidan~N. Gomez, \L{}ukasz Kaiser, and Illia Polosukhin.
\newblock Attention is all you need.
\newblock In \emph{NeurIPS}, page 6000–6010, 2017.

\bibitem[Wang et~al.(2019)Wang, Sun, Liu, Sarma, Bronstein, and Solomon]{wang2019dynamic}
Yue Wang, Yongbin Sun, Ziwei Liu, Sanjay~E. Sarma, Michael~M. Bronstein, and Justin~M. Solomon.
\newblock {Dynamic graph CNN for learning on point clouds}.
\newblock \emph{ACM TOG}, 38\penalty0 (5):\penalty0 146:1--146:12, 2019.

\bibitem[Wu et~al.(2015)Wu, Song, Khosla, Yu, Zhang, Tang, and Xiao]{wu2015}
Zhirong Wu, Shuran Song, Aditya Khosla, Fisher Yu, Linguang Zhang, Xiaoou Tang, and Jianxiong Xiao.
\newblock {3D ShapeNets: A deep representation for volumetric shapes}.
\newblock In \emph{CVPR}, pages 1912--1920, 2015.

\bibitem[Xiang et~al.(2021)Xiang, Zhang, Wei, and Gao]{xiang2021snowflakenet}
Xuhao Xiang, Chaoping Zhang, Zizhuang Wei, and Shenghua Gao.
\newblock {SnowflakeNet: Point cloud completion by snowflake point deconvolution with skip-transformer}.
\newblock In \emph{ICCV}, pages 5499--5509, 2021.

\bibitem[Xie et~al.(2020)Xie, Yao, Zhou, Mao, Zhang, and Sun]{xie2020grnet}
Haozhe Xie, Hongxun Yao, Shangchen Zhou, Jiageng Mao, Shengping Zhang, and Wenxiu Sun.
\newblock {GRNet}: Gridding residual network for dense point cloud completion.
\newblock In \emph{ECCV}, pages 365--381, 2020.

\bibitem[Yan et~al.(2022)Yan, Yan, Wang, Du, Wu, Xie, Pu, and Lu]{yan2022fbnet}
Xuejun Yan, Hongyu Yan, Jingjing Wang, Hang Du, Zhihong Wu, Di Xie, Shiliang Pu, and Li Lu.
\newblock {FBNet}: Feedback network for point cloud completion.
\newblock In \emph{ECCV}, pages 676--693, 2022.

\bibitem[Yang et~al.(2018)Yang, Feng, Shen, and Tian]{yang2018foldingnet}
Yaoqing Yang, Chen Feng, Yiru Shen, and Dong Tian.
\newblock {FoldingNet}: Point cloud auto-encoder via deep grid deformation.
\newblock In \emph{CVPR}, pages 206--215, 2018.

\bibitem[Yu et~al.(2022)Yu, Chen, Li, Sanghi, Shayani, Mahdavi-Amiri, and Zhang]{yu2022capri}
Fenggen Yu, Zhiqin Chen, Manyi Li, Aditya Sanghi, Hooman Shayani, Ali Mahdavi-Amiri, and Hao Zhang.
\newblock {CAPRI-Net}: Learning compact {CAD} shapes with adaptive primitive assembly.
\newblock In \emph{CVPR}, pages 11768--11778, 2022.

\bibitem[Yu et~al.(2018)Yu, Li, Fu, Cohen-Or, and Heng]{yu2018pu}
Lequan Yu, Xianzhi Li, Chi-Wing Fu, Daniel Cohen-Or, and Pheng-Ann Heng.
\newblock {PU-Net}: Point cloud upsampling network.
\newblock In \emph{CVPR}, pages 2790--2799, 2018.

\bibitem[Yu and Lafarge(2022)]{yu2022finding}
Mulin Yu and Florent Lafarge.
\newblock Finding good configurations of planar primitives in unorganized point clouds.
\newblock In \emph{CVPR}, pages 6367--6376, 2022.

\bibitem[Yu et~al.(2021)Yu, Rao, Wang, Liu, Lu, and Zhou]{yu2021pointr}
Xumin Yu, Yongming Rao, Ziyi Wang, Zuyan Liu, Jiwen Lu, and Jie Zhou.
\newblock {PoinTr}: Diverse point cloud completion with geometry-aware transformers.
\newblock In \emph{ICCV}, pages 12498--12507, 2021.

\bibitem[Yu et~al.(2023)Yu, Rao, Wang, Lu, and Zhou]{yu2023adapointr}
Xumin Yu, Yongming Rao, Ziyi Wang, Jiwen Lu, and Jie Zhou.
\newblock {AdaPoinTr}: Diverse point cloud completion with adaptive geometry-aware transformers.
\newblock In \emph{IEEE TPAMI}, pages 14114--14130, 2023.

\bibitem[Yuan et~al.(2018)Yuan, Khot, Held, Mertz, and Hebert]{yuan2018pcn}
Wentao Yuan, Tejas Khot, David Held, Christoph Mertz, and Martial Hebert.
\newblock {PCN}: Point completion network.
\newblock In \emph{2018 international conference on 3D vision (3DV)}, pages 728--737, 2018.

\bibitem[Zhao et~al.(2022)Zhao, Liu, Li, and Fu]{zhao2022seedformer}
Ruibin Zhao, Xinhai Liu, Jiancheng Li, and Hongbo Fu.
\newblock {SeedFormer: Patch seeds based point cloud completion with upsample transformer}.
\newblock In \emph{CVPR}, pages 11967--11976, 2022.

\bibitem[Zhu et~al.(2023)Zhu, Nan, Xie, Chen, Wang, Wei, and Qin]{zhu2023csdn}
Zhe Zhu, Liangliang Nan, Haoran Xie, Honghua Chen, Jun Wang, Mingqiang Wei, and Jing Qin.
\newblock {CSDN}: Cross-modal shape-transfer dual-refinement network for point cloud completion.
\newblock \emph{IEEE TVCG}, 2023.

\end{thebibliography}
}

\clearpage
\setcounter{page}{1}
\maketitlesupplementary

\renewcommand{\thefigure}{S\arabic{figure}}
\setcounter{figure}{0} 

\renewcommand{\thetable}{S\arabic{table}}
\setcounter{table}{0} 

\renewcommand{\theequation}{S\arabic{equation}}
\setcounter{equation}{0} 

Our supplementary materials provide visual demonstrations (\cref{sup:video}), instructions for reproducing our results (\cref{sup:reproducibility}), detailed implementation insights (\cref{sup:implementation}), and extended experimental analyses that augment the findings in the main paper (\cref{sup:analysis}).

\renewcommand\thesection{\Alph{section}}
\setcounter{section}{0}

\section{Video}
\label{sup:video}

The accompanying video, available on our project page\footnote{\label{projectpage}\url{https://parametric-completion.github.io}}, highlights the motivation behind our work, illustrates its key ideas, and demonstrates the reconstruction results.

\section{Reproducibility}
\label{sup:reproducibility}

The code and demo for our method are available in a public GitHub repository linked from the same project page\footref{projectpage}. Detailed instructions for setting up the environment and running the demo are provided in the \texttt{README.md} file.

\section{Implementation Details}
\label{sup:implementation}

\subsection{Dataset}

\begin{figure}[htb]
  \centering
  \includegraphics[width=0.95\linewidth]{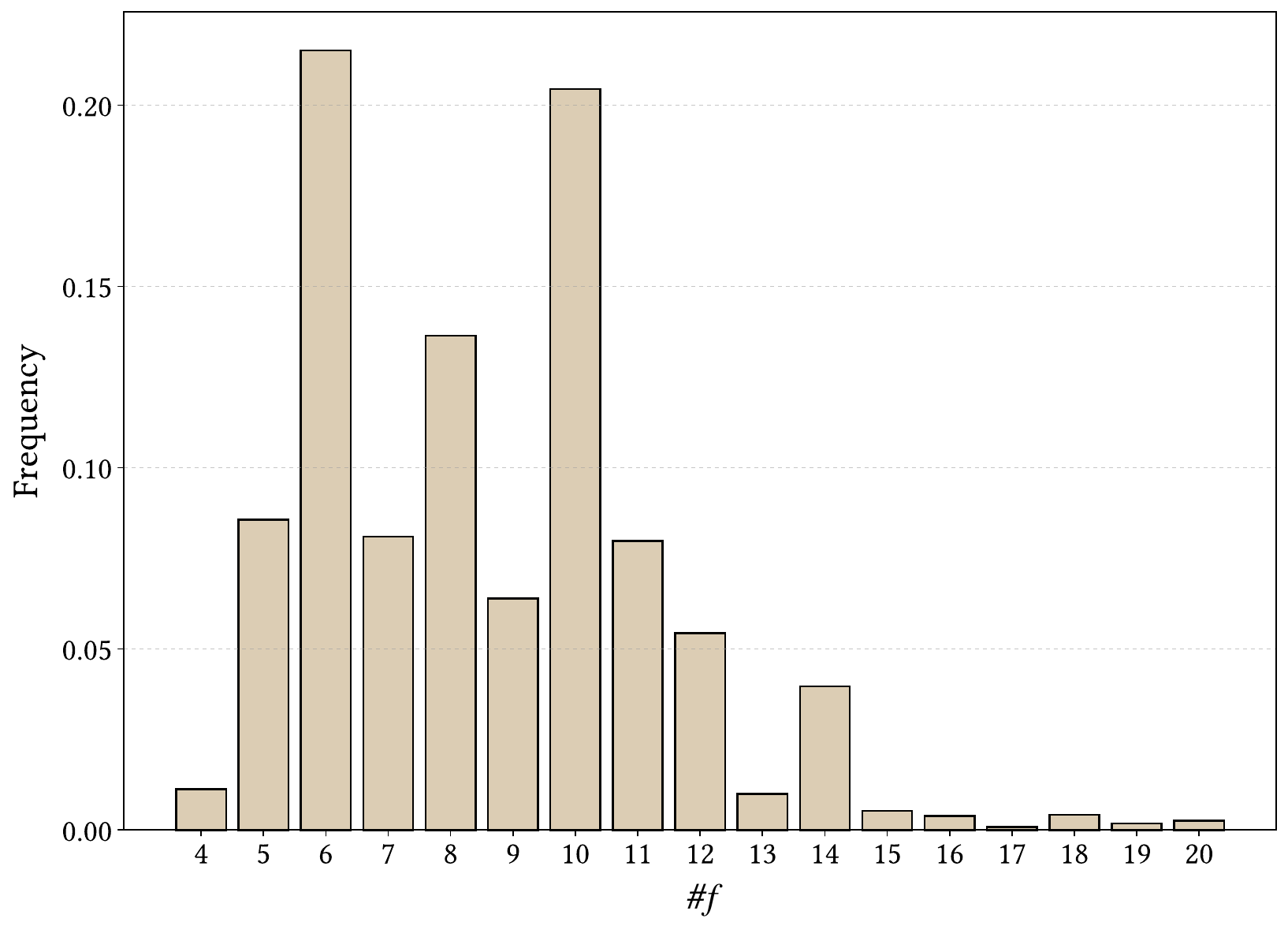}
  \caption{\textbf{Face count distribution in the dataset.}}
  \label{fig:dataset}
\end{figure}

\begin{figure}[htb]
  \centering
  \includegraphics[width=0.86\linewidth]{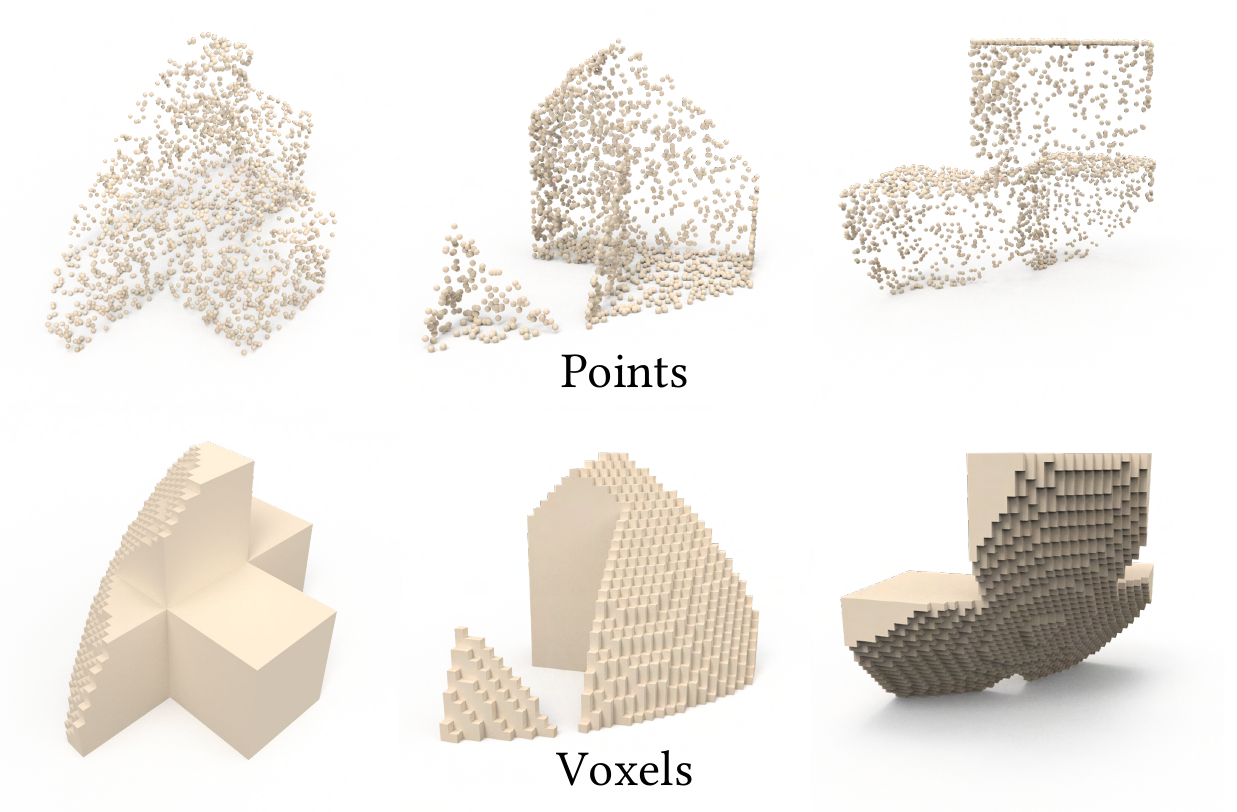}
  \caption{\textbf{Alignment between points and voxels.} This alignment ensures consistent representations of incomplete data.}
  \label{fig:alignment}
\end{figure}

We curated polygonal surface meshes from the ABC dataset~\cite{koch2019abc} to establish a consistent benchmark for evaluating polygonal surface reconstruction. We include only objects composed entirely of planar surfaces, excluding those with multiple parts or non-watertight geometries. The resulting dataset comprises 15,339 CAD models. The distribution of face counts is shown in \cref{fig:dataset}. Each input point cloud contains 2,048 points, while the ground truth point cloud comprises 8,192 points. 
For a fair comparison with voxel-based methods like BSP-Net~\cite{chen2020bspnet} and SECAD-Net~\cite{li2023secad} (\cref{sec:reconstruction}), the input point clouds are aligned with voxel representations. This alignment involves masking voxels based on their visibility from a selected viewpoint, excluding occluded voxels. \cref{fig:alignment} illustrates this alignment between point clouds and their corresponding voxel representation.

\subsection{Handling Failure Cases}

In cases of reconstruction failure (\eg, no geometry generated), metrics are calculated against a bounding box with a diagonal length of 1. This ensures a fair evaluation by penalizing failures in a controlled manner, mitigating potential bias. \cref{tab:sanity_eval} presents the results of a sanity check where all samples are treated as failures.

\begin{table}[htb]
  \centering
  \caption{\textbf{Sanity check results.} ``San.'' refers to cases where the normalized bounding box is used as the output geometry, ensuring unbiased metric calculation.}
  \footnotesize
  \begin{tabular}{lcccc}
    \toprule
    Method & CD $\downarrow$ & HD $\downarrow$ & NC $\uparrow$   \\
    \midrule
    San. & 16.36 & 22.06 & 0.550 \\ 
    PaCo (ours) & 1.87 & 4.09 & 0.943 \\
    \bottomrule
  \end{tabular}
  \label{tab:sanity_eval}
\end{table}

\subsection{Hyperparameters}

PaCo is implemented in PyTorch and optimized using the AdamW optimizer with an initial learning rate of $10^{-4}$, a weight decay of $5 \times 10^{-4}$, and a learning rate decay of 0.9 every 20 epochs. The encoder and decoder depths are set to 8 and 12, respectively. For ablation studies (\cref{sec:ablation}), the encoder depth is set to 6 and the decoder depth to 8 to reduce computational complexity. The encoder produces 128 point proxies, and the number of plane proxies \( K \) after padding is set to 20. A total of 40 queries (\( M \)) are used. For the loss terms in \cref{eq:loss}, we empirically set \( \beta_2 = \beta_4 = 20 \), and \( \beta_3 = 2 \). To address class imbalance, we set
\begin{equation}
\beta_1 =
\begin{cases}
0.4, & \text{if } c_i = \emptyset, \\
1, & \text{otherwise.}
\end{cases}
\end{equation}
All the competing methods~\cite{yang2018foldingnet, yuan2018pcn, xie2020grnet, yu2021pointr, yu2023adapointr, cai2024orthogonal, chen2020bspnet, li2023secad, garland1997qem, chen2023robust} and reconstruction solvers~\cite{nan2017polyfit, bauchet2020kinetic, sulzer2024concise} are used with their default settings.

\section{Additional Analysis}
\label{sup:analysis}

\subsection{Evaluation on Primitive Parameters}

In the main paper, we use normal consistency (NC) as one of the metrics to evaluate the quality of the reconstruction results. The NC is calculated from surface-sampled points and is an indirect indicator of the accuracy of the primitive parameters recovered by parametric completion.
To complement the indirect surface-based evaluation (see \cref{sec:metrics}), we introduce a direct metric, $\text{NC}_\mathrm{prim}$, defined as the average normal consistency between the predicted and ground truth primitives,  to quantify the accuracy of the recovered primitive parameters. For the competing methods, GoCoPP~\cite{yu2022finding} is used to extract primitives from the completed points. Our method achieves outstanding performance, demonstrating high-quality primitives with an $\text{NC}_\mathrm{prim}$ of 0.976.

\begin{table}[htb]
  \centering
  \caption{\textbf{Evaluation on primitive normal consistency.} PaCo produces planar primitives with the highest $\text{NC}_\mathrm{prim}$.}
  \footnotesize
  \begin{tabular}{lc}
    \toprule
    Method  & $\text{NC}_\mathrm{prim}$ $\uparrow$ \\
    \midrule
    PCN \cite{yuan2018pcn} & 0.605 \\
    FoldingNet \cite{yang2018foldingnet} & 0.849 \\
    GRNet \cite{xie2020grnet} & 0.752 \\
    PoinTr \cite{yu2021pointr} & 0.863 \\
    AdaPoinTr \cite{yu2023adapointr} & 0.930 \\
    ODGNet \cite{cai2024orthogonal} & 0.933 \\
    PaCo (ours) & \textbf{0.976} \\
    \bottomrule
  \end{tabular}
  \label{tab:parametric}
\end{table}

\subsection{Surface Complexity}

\begin{figure}[htb]
  \centering
  \includegraphics[width=0.92\linewidth]{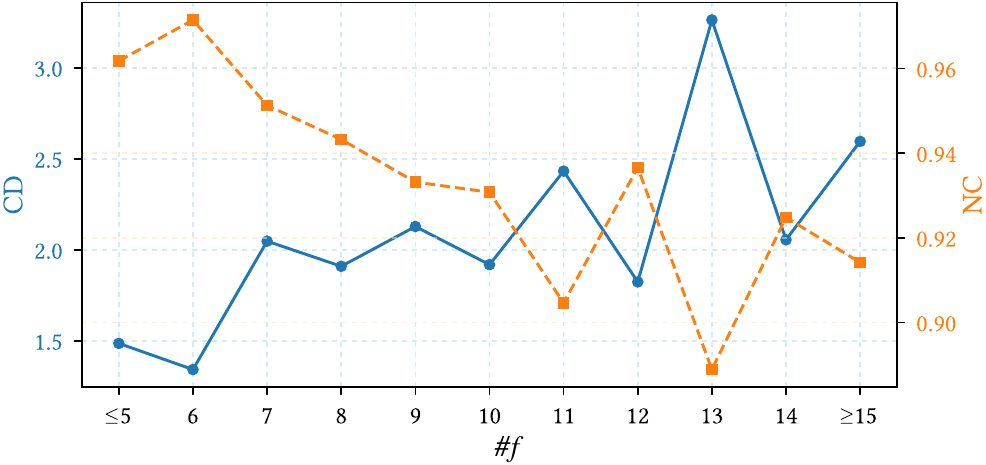}
  \caption{\textbf{Performance concerning surface complexity.} As the complexity increases, CD tends to rise while NC declines.}
  \label{fig:complexity}
\end{figure}

\cref{fig:complexity} presents the performance of our method with respect to surface complexity. As the geometric complexity increases, CD tends to rise while NC declines. This trend arises from the inherent challenges of learning complex structures and the underrepresentation of highly complex samples in the dataset (see \cref{fig:dataset}).

\subsection{Point Density}
\label{sup:npoints}

Since competing methods often rely on point density for improved metrics, we normalize all outputs by down-sampling or up-sampling to a consistent total of 8,192 points. This ensures a fair comparison and prevents metric variations from being attributed to point count differences. \cref{tab:npoints} shows that this normalization has a negligible impact on performance for PaCo, demonstrating its robustness in recovering accurate parametric representations independent of point density.

\begin{table}[htb]
  \centering
  \caption{\textbf{Impact of number of points for PaCo.} ``S'' indicates sampling to 8,192 points. The number of points has no significant impact on the reconstruction.}
  \footnotesize
  \begin{tabular}{lc|ccccc}
    \toprule
    Solver & S & CD $\downarrow$ & HD $\downarrow$ & NC $\uparrow$ & FR $\downarrow$ \\
    \midrule
    \multirow{2}{*}{KSR \cite{bauchet2020kinetic}} &  & 1.91 & 4.14 & 0.940 & 0.25 \\
    & \checkmark & 1.92 & 4.19 & 0.945 & 0.28 \\
    \midrule
    \multirow{2}{*}{COMPOD \cite{sulzer2024concise}} & & 1.94 & 4.42 & 0.940 & 0.25 \\
    & \checkmark & 1.95 & 4.56 & 0.943 & 0.34 \\
    \midrule
    \multirow{2}{*}{PolyFit \cite{nan2017polyfit}} & & 1.87 & 4.09  & 0.943 & 0.48 \\
    & \checkmark & 1.93 & 4.48  & 0.941 & 0.63 \\
    \bottomrule
  \end{tabular}
  \label{tab:npoints}
\end{table}

\subsection{Number of Proxies}

We set the number of proxies to be greater than the maximum number of faces of the samples in the dataset. This surplus allows the primitive selector to distinguish between positive and negative primitives more effectively. \cref{tab:nproxy} presents the results using different numbers of plane proxies, where 40 proxies achieve the best performance. The optimal number of proxies is likely correlated to the distribution of face counts (see \cref{fig:dataset}).

\begin{table}[htb]
    \centering
    \caption{\textbf{Performance with different proxy numbers.} Our method performs the best with 40 proxies.}
    \footnotesize
    \begin{tabular}{ccccc}
        \toprule
        Num. Proxies & CD $\downarrow$ & HD $\downarrow$ & NC $\uparrow$ \\
        \midrule
        30 & 2.75 & 6.98 & 0.906 \\
        40 & \textbf{2.11} & \textbf{5.26} & \textbf{0.941} \\
        50 & 2.64 & 6.86 & 0.908 \\
        \bottomrule
    \end{tabular}
    \label{tab:nproxy}
\end{table}

\subsection{Evaluation on Real Data}

\begin{figure}[ht]
  \centering
  \includegraphics[width=0.99\linewidth]{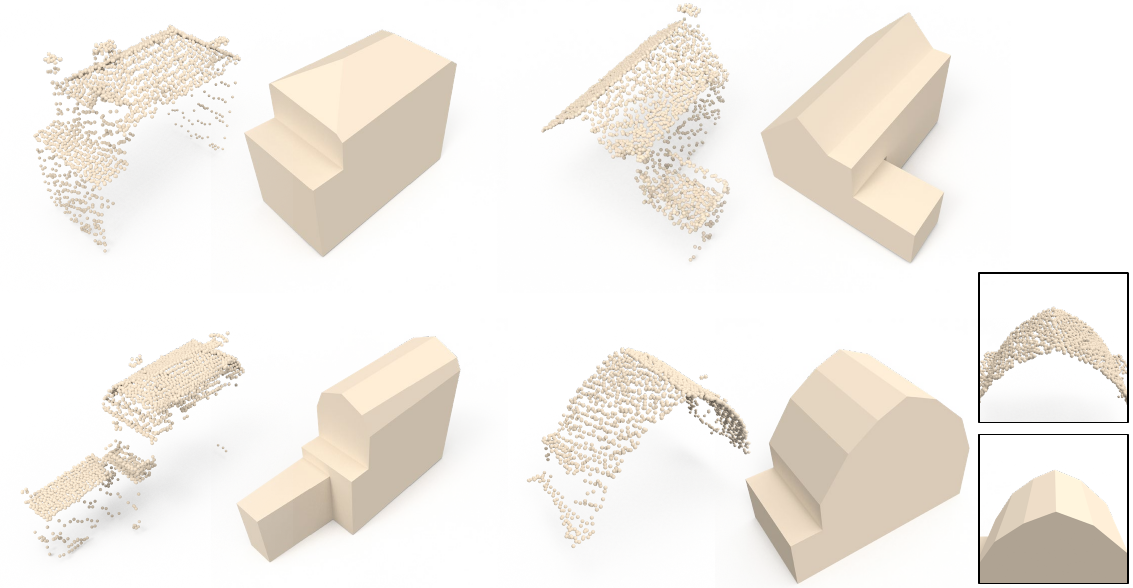}
  \caption{\textbf{Buildings reconstructed from real-world airborne LiDAR data with noise and outliers.} The fine-tuned PaCo model generalizes to unseen airborne LiDAR data.}
  \label{fig:realworld}
\end{figure}

We evaluate PaCo on real airborne LiDAR point clouds for building reconstruction to assess its transferability (\cref{fig:realworld}). Although PaCo was trained on the ABC dataset~\cite{koch2019abc} and fine-tuned with only 2,000 airborne LiDAR samples\footnote{\url{https://www.ahn.nl/}}, it successfully reconstructs heavily occluded structures without relying on explicit rules. This demonstrates its robustness to noise and outliers and, notably, its ability to approximate freeform surfaces using planar primitives.

\subsection{Inference Speed}

\begin{figure}[ht]
  \centering
  \includegraphics[width=0.90\linewidth]{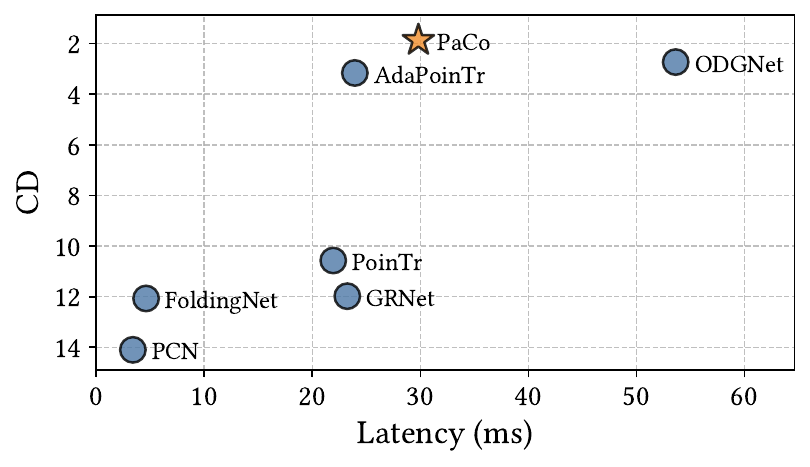}
  \caption{\textbf{Chamfer distance (CD) vs. inference latency.} CD is computed on meshes via the PolyFit solver, and latency is measured on an NVIDIA A40 GPU (excluding I/O).}
  \label{fig:efficiency}
\end{figure}

\cref{fig:efficiency} presents the inference latency comparisons across different completion methods. PaCo achieves an average inference time of 29.8\,ms, nearly twice as fast as the strongest competitor, ODGNet (53.6\,ms), while reducing CD by 32\%. Notably, bipartite matching applies only during training and does not affect inference.

\end{document}